%% file: main.tex
\documentclass[11pt, DIV=17, abstract, headings=normal]{scrartcl}

\usepackage[sc]{mathpazo}
\usepackage[T1]{fontenc}

\usepackage{etoolbox}
\AtBeginEnvironment{abstract}{\footnotesize}

\usepackage{etoolbox}

\setkomafont{sectioning}{\scshape}
\setkomafont{author}{\small}

\usepackage{microtype}
\usepackage{graphicx}
\usepackage{booktabs} % for professional tables
\usepackage{caption}
\usepackage{subcaption} % for subcaption boxes
\usepackage{amsmath}
\usepackage{bm}      % for bold greek letters
\usepackage{amssymb} % more math symbols
\usepackage{amsthm}
\usepackage{bbm}  % blackboard 1
\usepackage{hyperref}
\usepackage{paralist}
\usepackage{wrapfig}
\usepackage{siunitx}
\usepackage{etoolbox}

\usepackage{comment}
\usepackage[inline]{enumitem}
\usepackage[sort, numbers, compress]{natbib}
\usepackage{multirow} 
\PassOptionsToPackage{sort, numbers, compress}{natbib}

\setcitestyle{square}
\renewcommand{\subsubsection}[1]{\textbf{#1}

} % newlines are deliberate

\DeclareMathOperator*{\argmin}{arg\,min}

\newcommand{\reals}{\mathbb{R}}
\newcommand{\naturals}{\mathbb{N}}
\newcommand{\sig}{\mathrm{Sig}^N}
\newcommand{\dataspace}{\mathcal{X}}
\newcommand{\lspace}{\mathcal{Y}}
\newcommand{\seriesspace}{\mathcal{S}}
\newtheorem{theorem}{Theorem}
\makeatletter
\patchcmd{\@maketitle}{\titlefont\huge}{\titlefont\small}{}{}
\makeatother

\title{\LARGE Path Imputation Strategies for Signature Models of Irregular Time Series}

\author{
  Michael Moor$^{1,2}$, Max Horn$^{1,2}$, Christian Bock$^{1,2}$, Karsten Borgwardt$^{1,2}$, Bastian Rieck$^{1,2}$\\[0.5cm]
  \footnotesize\textsc{$^{1}$Department of Biosystems Science and Engineering, ETH Zurich, Switzerland}\\
  \footnotesize\textsc{$^{2}$SIB Swiss Institute of Bioinformatics, Switzerland}\\
  \footnotesize\texttt{\{firstname.lastname@bsse.ethz.ch\}}
}

\date{}

\begin{document}
\maketitle

\begin{abstract}
\noindent The signature transform is a `universal nonlinearity' on the space of
continuous vector-valued paths, and has received attention for use in
machine learning on time series. However, real-world temporal data is typically observed at discrete points in time, and must first be transformed into a continuous path before
signature techniques can be applied. We make this step explicit by characterising it as an imputation
problem, and empirically assess the impact of various imputation strategies when
applying signature-based neural nets to irregular time series data. For one of these
strategies, Gaussian process~(GP) adapters, we propose an
extension~(GP-PoM) that makes uncertainty information directly available
to the subsequent classifier while at the same time preventing costly
Monte-Carlo~(MC) sampling. In our experiments, we find that the choice
of imputation  drastically affects shallow signature models, whereas
deeper architectures are more robust. Next, we observe that
uncertainty-aware \emph{predictions}~(based on GP-PoM or indicator
imputations) are beneficial for predictive performance, even compared to
the uncertainty-aware \emph{training} of conventional GP adapters.
In conclusion, we have demonstrated that the path construction is indeed
crucial for signature models and that our proposed strategy leads to
competitive performance in general, while improving robustness of
signature models in particular.
\end{abstract} 

%%%%%%%%%%%%%%%%%%%%%%%%%%%%%%%%%%%%%%%%%%%%%%%%%%%%%%%%%%%%%%%%%%%%%%%%
\section{Introduction}
%%%%%%%%%%%%%%%%%%%%%%%%%%%%%%%%%%%%%%%%%%%%%%%%%%%%%%%%%%%%%%%%%%%%%%%%

Originally described by \citet{Chen54, Chen57, Chen58} and popularised
in the theory of rough paths and controlled differential
equations~\cite{lyons1998differential, FritzVictoir10, lyons2014rough},
the \emph{signature transform}, also known as the \emph{path signature}
or simply \emph{signature}, acts on a continuous vector-valued path of
bounded variation, and returns a graded sequence of statistics, which
determine a path up to a negligible equivalence class. Moreover,
\emph{every} continuous function of a path can be recovered by applying
a linear transform to this collection of statistics~\citep[Proposition
A.6]{kidger2019deep}.
This `universal nonlinearity' property makes the signature a promising nonparametric
feature extractor in both generative and
discriminative learning scenarios.
Further properties include the signature's uniqueness~\citep{hambly2010uniqueness}, as well as factorial decay of its higher order terms~\citep{lyons1998differential}. These theoretical foundations have been accompanied by outstanding empirical results when applying signatures to clinical time series classification tasks~\citep{reyna2019early, morrill2019signature}.

Due to their similarities, we may hope that tools that
apply to continuous paths can \emph{also} be applied to multivariate
time series. But since multivariate time series are not continuous
paths, one first needs to construct a continuous path before signature techniques are applicable.
Previous work~\citep{levin2013, kidger2019deep, fermanian2019embedding}
characterised this construction as an embedding problem, and
typically considered it a minor technical detail.
This is exacerbated by the---perfectly sensible---behaviour of software
for computing the signature~\citep{iisignature, signatory}, which
commonly considers a continuous piecewise linear path as an input,
described by its sequence of knots, i.e.\ values.
Since such sequences resemble a sequence of data, the signature is
sometimes interpreted as operating on sequences of data rather than on
paths~\cite{kidger2019deep, levin2013}.
By contrast, here we show that considering the path construction
process is crucial for achieving competitive predictive
performance: we reinterpret the task of constructing a continuous path,
turning it from an embedding problem to an imputation problem, which we
call \emph{path imputation}.

While previous research concerning the signature transform
focused on its excellent theoretical properties, such as sampling
independence~\citep[Proposition A.7]{kidger2019deep}, our findings show
that this does not
necessarily correspond to empirical performance.
We perform a thorough investigation of multiple imputation schemes in combination with various models that can potentially employ signatures.
Furthermore, motivated by the fact that missingness itself can be
informative for time series classification~\citep{rubin1976inference}, we propose a novel imputation strategy: an extension of Gaussian process adapters~\citep{li2016scalable, futoma2017mgp}, which exploits uncertainty information during each prediction step and which is beneficial for signature models, but also of independent interest.
We make our code available under \url{https://osf.io/bg9cw/?view_only=5193e93118d84a5f9be4f261df4c0a06}.

%%%%%%%%%%%%%%%%%%%%%%%%%%%%%%%%%%%%%%%%%%%%%%%%%%%%%%%%%%%%%%%%%%%%%%%%
\section{Related work}
%%%%%%%%%%%%%%%%%%%%%%%%%%%%%%%%%%%%%%%%%%%%%%%%%%%%%%%%%%%%%%%%%%%%%%%%

A key motivation for this work is the use of the signature transform in
machine learning: recent work~\citep{primer2016,
kormilitzlin2016, yang2016rotation, li2017lpsnet, yang2017leveraging,
PerezArribas2018, morrill2019sepsis} typically employed the signature transform as a nonparametric feature extractor, on top of which
a model is learnt. A growing body of work has also investigated how to
integrate the signature transform more tightly with neural networks; 
\citet{jeremythesis}, \citet{logsigrnn}, and \citet{kidger2019deep} all
study how to use the signature transform~(or variants thereof) within typical neural network models.
\citet{chevyrev2018signature, kiraly2019kernels} study how the signature
transform may be used to define a \emph{kernel}---i.e.\ a symmetric,
positive definite function that is typically used as a similarity
measure---on path space, while \citet{toth2019gp} show how this kernel
may be used to define a Gaussian process.
In much of this work, data has been converted into a continuous path via
linear interpolation. Some authors~\citep{primer2016,
fermanian2019embedding} have additionally considered `rectilinear'
interpolation, which is similar. \citet{levin2013} present the
`time-joined transformation', which is a hybrid of the two, such that
the resulting path exhibits a causal dependence on the data.  However,
to our knowledge, no prior work has regarded (and empirically investigated) this as an imputation
problem.

\paragraph{Imputation schemes} The general problem of imputing data is well-known and well-studied, and
we will not attempt to describe it here; see for example \citet[Chapter 25]{gelman2007dataanalysis}.
Imputation methods typically only fill in missing discrete data
points, and do not attempt to impute the underlying continuous path.
Gaussian process adapters~\citep{li2016scalable}, by contrast, are
capable of imputing a \emph{full} continuous path, from which we
may sample arbitrarily. Hence, this framework will be
considered more closely in this paper.
We note that there are also other approaches that
perform imputation end-to-end with a downstream classifier~\citep{shukla2018interpolationprediction}
and methods that skip the imputation step altogether based on recently-proposed Neural-ODE like architectures~\citep{rubanova2019latent, kidger2020neuralcde}, variants of recurrent neural networks \cite{che2018recurrent} or set functions~\cite{horn2019set}.
However, the scope of this work is to specifically assess the impact of path imputations for the signature, hence we deem the larger comparison including imputation-free scenarios interesting for future work, while it bypasses the central point of this paper.

%%%%%%%%%%%%%%%%%%%%%%%%%%%%%%%%%%%%%%%%%%%%%%%%%%%%%%%%%%%%%%%%%%%%%%%%
\section{Background: Signature transform and Gaussian process adapters} % 
%%%%%%%%%%%%%%%%%%%%%%%%%%%%%%%%%%%%%%%%%%%%%%%%%%%%%%%%%%%%%%%%%%%%%%%%
\paragraph{Path signatures}
Let $f = (f_1, \dots, f_d) \colon [a, b] \to \reals^d$ be a continuous, piecewise
differentiable path. Then the \emph{signature transform up to depth $N$}
is
\begin{equation}\label{eq:signature}
    \sig(f)=\left(\left(\underset{\,a<t_{1}<\cdots<t_{k}<b}{\int \cdots \int} \prod_{j=1}^{k} \frac{\mathrm{d} f_{i_{j}}}{\mathrm{d} t}\left(t_{j}\right) \mathrm{d} t_{1} \cdots \mathrm{d} t_{k}\right)_{1 \leq i_{1}, \ldots, i_{k} \leq d}\right)_{1 \leq k \leq N}.
\end{equation}

\begin{wrapfigure}{3}{5cm}
    \centering
    \vspace{-1em}
	\includegraphics[width=0.25\columnwidth]{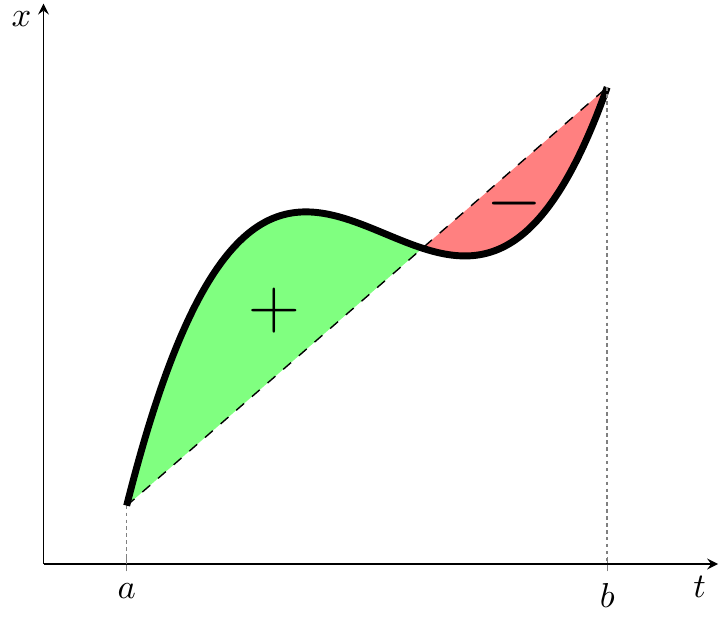}
	\caption{Given a path (bold), its Levy area is its signed area with respect to the chord joining its endpoints.}\label{fig:sig_path}
	\vspace{-2em}
\end{wrapfigure}

This definition can be extended to paths of bounded variation by
replacing these integrals with Stieltjes integrals with respect to each
$f_{i_j}$.
In brief, the signature transform may be interpreted as extracting
information about \emph{order} and \emph{area} of a path.
One may interpret its terms as `the area/order of one channel with
respect to some collection of other channels'. To give an explicit example: first level terms simply describe the increment of the path with respect to one channel, whereas second-order terms are related to the \emph{Levy area} of the path, as shown for a one-dimensional example in Figure~\ref{fig:sig_path}.

For an exposition on the properties of the signature transform and its use in machine learning, please refer to \citet{primer2016} or \citet[Appendix A]{kidger2019deep}. For building more intuition, in Section~\ref{sec: comparison fourier} of the appendix, we compare the signature to more well-known transforms.

\paragraph{Computing the signature transform} \label{sec: computing sig}
Continuous piecewise linear paths are the paths of choice, computationally speaking, due to the fact that this is the only case for which efficient algorithms for computing the signature transform are known \cite{signatory}.

This is not a serious hurdle when one wishes to compute the signature of a path $f$ that is not piecewise linear---as the signature of piecewise linear approximations to $f$ will tend towards the signature of $f$ as the quality of the approximation increases---but it does enforce this requirement on our imputation schemes.
Thus, all of the imputation schemes we examine will first seek to select a collection of points in data space (not necessarily only where we had data before), and for computing the signature we join them up into a piecewise linear path.

%%%%%%%%%%%%%%%%%%%%%%%%%%%%%%%%%%%%%%%%%%%%%%%%%%%%%%%%%%%%%%%%%%%%%%%%
\paragraph{Notation}
%%%%%%%%%%%%%%%%%%%%%%%%%%%%%%%%%%%%%%%%%%%%%%%%%%%%%%%%%%%%%%%%%%%%%%%%
%
We define the space of time series over a set $A$ by
\begin{equation}
    \seriesspace(A) = \{((t_1, x_1), \ldots, (t_n, x_n)) \,\vert\, t_i \in \reals, x_i \in A, n \in \naturals, \text{ such that } t_1 \leq \cdots \leq t_n\}.\label{eq:seriesspace}
\end{equation}

Furthermore, let $\lspace$ be a set and let $\dataspace_j = \reals$ for $j \in \{1,
\ldots, d\}$ and $d \in \naturals$. Then we assume that we observe
a dataset of labelled time series $(\mathbf{x}_k, y_k)$ for $k \in \{1,
\ldots, N\}$, where $\mathbf{x}_k \in \seriesspace(\dataspace^*)$ and
$y_k \in \lspace$, with $\dataspace^* = \prod_{j = 1}^d(\dataspace_j
\cup \{*\})$ and $*$ representing no observation.
We similarly define $\dataspace = \prod_{j = 1}^d\dataspace_j$. Thus,
$\dataspace$ is the data space, while $\dataspace^*$ is the data space
allowing missing data, and $\lspace$ is the set of labels.

%%%%%%%%%%%%%%%%%%%%%%%%%%%%%%%%%%%%%%%%%%%%%%%%%%%%%%%%%%%%%%%%%%%%%%%%
\paragraph{Gaussian process adapter} \label{sec: GPadapter}
%%%%%%%%%%%%%%%%%%%%%%%%%%%%%%%%%%%%%%%%%%%%%%%%%%%%%%%%%%%%%%%%%%%%%%%%
%
Some of the imputation schemes we consider are based on
the uncertainty aware-framework of multi-task Gaussian process
adapters~\citep{li2016scalable, futoma2017mgp}.
Let $\mathcal{W}, \mathcal{H}$ be some sets. Let $\ell \colon \lspace
\times \lspace \to [0, \infty)$ be a loss function. Let $F \colon
\dataspace^{[a, b]} \times \mathcal{W} \to \lspace$, be some (typically
neural network) model, with $\mathcal{W}$ interpreted as a space of
parameters. Let \begin{align*} \mu \colon [a, b] \times
\seriesspace(\dataspace^*) \times \mathcal{H} &\to \dataspace\\ \Sigma
\colon [a, b] \times [a, b] \times \seriesspace(\dataspace^*) \times
\mathcal{H} &\to \dataspace    \end{align*} be mean and covariance
functions, with $\mathcal{H}$ interpreted as a space of
hyperparameters. The dependence on $\seriesspace(\dataspace^*)$ is used to
represent conditioning on observed values.

Then the goal is to solve
\begin{equation}\label{eq:gp-mc}
\argmin_{\mathbf{w} \in \mathcal{W},\bm{\eta} \in \mathcal{H}} \sum_{k=1}^N \overbrace{\rule{0pt}{0.5cm} \mathbb{E}_{\mathbf{z}_k \sim \mathcal{N}\left( \mu(\cdot, \mathbf{x}_k, \eta), \Sigma(\cdot, \cdot, \mathbf{x}_k, \eta)\right) } \big[ \ell(F(\mathbf{z}_k, \mathbf{w}),
y_k) \big] }^{\text{$E_k$}}.
\end{equation}
As this expectation is typically not tractable, it is estimated by MC
sampling with $S$ samples, i.e.\
\begin{equation}
E_k \approx \frac{1}{S} \sum_{s=1}^{S} \ell(F(\mathbf{z}_{s, k}, \mathbf{w}), y_k),
\end{equation}
where
\begin{equation}
    \mathbf{z}_{s, k} \sim \mathcal{N}\left( \mu(\,\cdot\,, \mathbf{x}_k, \eta), \Sigma(\,\cdot\,, \,\cdot\,, \mathbf{x}_k, \eta)\right).
\end{equation}
Alternatively, one may forgo allowing the uncertainty to propagate through $F$ by instead passing the posterior mean directly to $F$; this corresponds to solving
\begin{equation}\label{eq:gp-mean}
  \argmin_{\mathbf{w} \in \mathcal{W},\bm{\eta} \in \mathcal{H}} \sum_{k=1}^N \ell(F(\mu(\,\cdot\,,\mathbf{x}_k, \eta), \mathbf{w}), y_k).
\end{equation}

%%%%%%%%%%%%%%%%%%%%%%%%%%%%%%%%%%%%%%%%%%%%%%%%%%%%%%%%%%%%%%%%%%%%%%%%
\section{Path imputations for signature models}
%%%%%%%%%%%%%%%%%%%%%%%%%%%%%%%%%%%%%%%%%%%%%%%%%%%%%%%%%%%%%%%%%%%%%%%%

Signatures act on continuous paths. However, in real-world
applications, temporal data typically appears as a discretised
collection of measurements, potentially irregularly-spaced and
asynchronously observed. To apply the signature to this data,
it first has to be converted into a continuous path. 
We believe this step to have a significant impact on the
resulting signature, and thus also on models employing the signature.
To assess this hypothesis, we explicitly treat this transformation
as a \emph{path imputation}, i.e.\ a mapping of the form $ \phi \colon \seriesspace(\dataspace^*) \to (\reals \times \dataspace)^{[a, b]}.$
\paragraph{Task} We aim to learn a function $ g \colon \seriesspace(\dataspace^*) \to
\lspace$, which decomposes to $g = F \circ \phi$, where $F$ refers to
a classifier, mapping from $(\reals \times \dataspace)^{[a, b]} \times \mathcal{W} $ to
$\lspace$. Given a loss function $\ell$ and a set of $p$ path
imputation strategies, ${\Phi} =(\phi_i)_{i=1}^p$, we seek to minimise the
objective:
\begin{equation}
    \argmin_{\phi_i \in \Phi, \mathbf{w} \in \mathcal{W}} \quad \mathbb{E}_{(\mathbf{x},y) \sim P(\seriesspace(\dataspace^*),\lspace) } \left[ \ell( g(\mathbf{x}; \phi_i, \mathbf{w}), y) \right]
    \label{eq:objective}
\end{equation} 
Even though Equation~\eqref{eq:objective} could be formulated more
\emph{implicitly}~(i.e.\ without any explicit imputation step),
this formulation enables us to make explicit how the signature transform
`interprets' the raw data for downstream classification tasks. We further motivate this need for assessing the path construction in Section~\ref{sec:Fragile dependence} by showing that a single imputed value can affect the Levy area which is computed with the signature. 

\paragraph{Path imputation strategies}
For our analysis, we consider the following set of strategies for path imputation, namely
\begin{inparaenum}[(1)]
    \item linear interpolation,
    \item forward filling,
    \item indicator imputation,
    \item zero imputation,
    \item causal imputation\footnote{This strategy is similar to the
      time-joined transformation \citep{levin2013}. For more details,
    please refer to Section \ref{sec:Causal signature imputation} in the
  appendix.}, and
    \item Gaussian process adapters~(GP).
\end{inparaenum}
Strategies 1--5 can be seen as a fixed preprocessing step, whereas GP
adapters~(strategy~6) are optimised end-to-end with the downstream task. For more
details regarding these strategies, please refer to Section~\ref{supp: Imputation} in the appendix. As indicated in Section~\ref{sec: computing sig}, for computing the signature efficiently~(i.e.\ computed in terms of standard tensor operations \citep[Proposition A.3]{kidger2019deep}), the imputed time series are transformed into piecewise linear paths beforehand.

\begin{figure}[t]
    \centering
    \hspace*{1cm}
    \includegraphics[width=0.7\columnwidth]{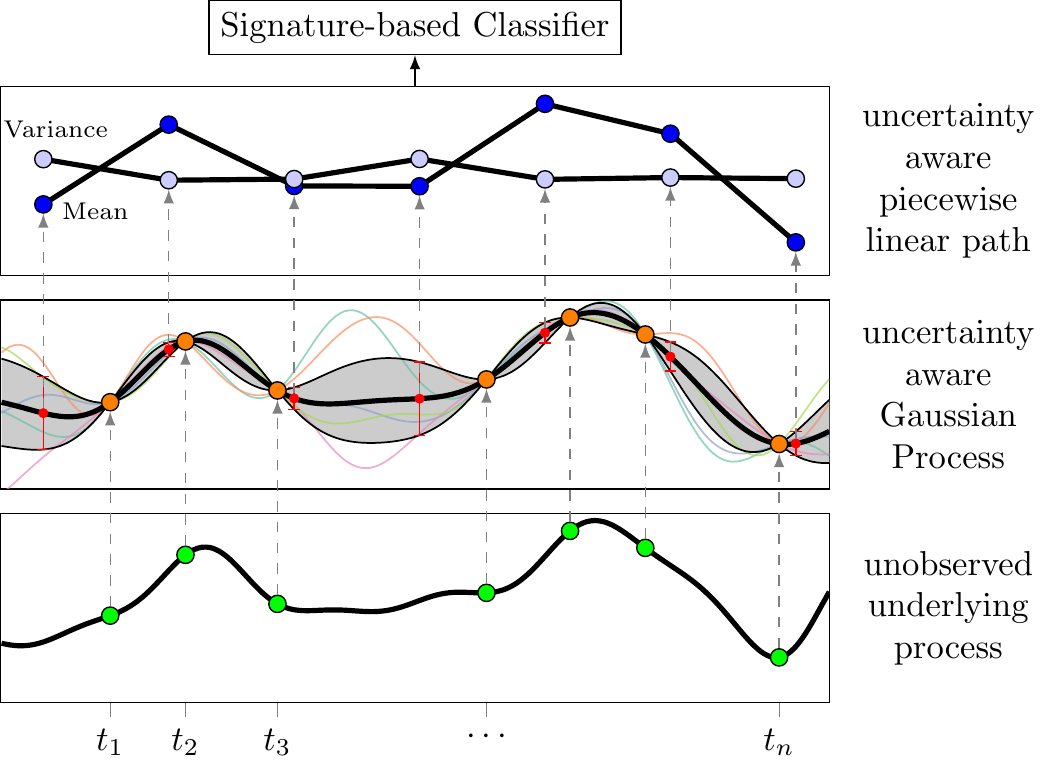}
    \caption{%
      Overview of our proposed extension of GP adapters, GP-PoM,
      leveraging both posterior moments~(mean and variance). In
      comparison, the conventional GP adapter feeds MC samples~(faded
      colours in the background) drawn from the GP posterior into the
      classifier.
    }
    \label{fig:overview}
\end{figure}

\paragraph{GP adapter with posterior moments}
For conventional GP adapters, one major drawback with the formulations
of \citet{li2016scalable} and \citet{futoma2017mgp}, as described in
Equation~\eqref{eq:gp-mc}, is that approximating the expectation outside
of the loss function with MC sampling is expensive.
During prediction, \citet{li2016scalable} proposed to overcome this
issue by sacrificing the uncertainty in the loss function and to simply
pass the posterior mean, as in Equation~\eqref{eq:gp-mean}\footnote{%
  Equations \eqref{eq:gp-mc} and \eqref{eq:gp-mean} are of course not in
  general equal, so following \citet{futoma2017mgp}, our standard GP
  adapter uses MC sampling both in training and testing.
}.
To address both points, we propose to instead also pass the posterior
covariance of the Gaussian process to the classifier $F$. This saves the
cost of MC sampling whilst explicitly providing~$F$ with uncertainty
information during the prediction\footnote{%
Even if MC sampling
is used during prediction, $F$ has no per-sample access to
uncertainty about the imputation.
}.
However, the full covariance matrix may become very large, and it is
not obvious that all interactions are relevant to the subsequent
classifier.
This is  why we simplify matters by taking the posterior
\emph{variance} at every point, and concatenate it with the posterior
mean at every point, to produce a path whose evolution also captures the
uncertainty at every point:
\begin{align}
    \tau &\colon [a, b] \times \seriesspace(\dataspace^*) \times \mathcal{H} \to \dataspace \times \dataspace\\
    \tau &\colon t, \mathbf{x}, \eta \mapsto (\mu(t, \mathbf{x}, \eta), \Sigma(t, t, \mathbf{x}, \eta)).
\end{align}
This corresponds to solving
\begin{equation}\label{eq:gp-moments}
\argmin_{\mathbf{w} \in \mathcal{W},\bm{\eta} \in \mathcal{H}} \sum_{k=1}^N \ell(F(\tau(\,\cdot\,,\mathbf{x}_k, \eta), \mathbf{w}), y_k),
\end{equation}
where instead now $F \colon (\dataspace \times \dataspace)^{[a, b]}
\times \mathcal{W} \to \lspace$.
As this approach leverages information from both posterior moments (mean and variance), we refer to it as posterior moments GP adapter, or short \textsc{GP-PoM}.
Figure~\ref{fig:overview} gives an overview of \textsc{GP-PoM}. 
In our context of interest, when $F$ is a signature model, it is now
straightforward to compute the signature of the Gaussian process, simply
by querying many points to construct a piecewise linear approximation to
the process.  The choice of kernel has non-trivial mathematical
implications for this procedure: for example if a Mat{\'e}rn 1/2 kernel
is chosen, then the resulting path is not of bounded variation and the
definition of the signature transform given in Equation~\eqref{eq:signature}
does not hold, and rough path theory~\citep{lyons1998differential} must instead be invoked to define the
signature transform. However, in this work we use RBF kernels, and therefore, this caveat does not apply to our case.

%%%%%%%%%%%%%%%%%%%%%%%%%%%%%%%%%%%%%%%%%%%%%%%%%%%%%%%%%%%%%%%%%%%%%%%%
\section{Experiments}
%%%%%%%%%%%%%%%%%%%%%%%%%%%%%%%%%%%%%%%%%%%%%%%%%%%%%%%%%%%%%%%%%%%%%%%%

We first introduce our experimental setup~(datasets and model
architectures) before presenting and discussing quantitative results.

%%%%%%%%%%%%%%%%%%%%%%%%%%%%%%%%%%%%%%%%%%%%%%%%%%%%%%%%%%%%%%%%%%%%%%%%
\paragraph{Datasets and preprocessing}
%%%%%%%%%%%%%%%%%%%%%%%%%%%%%%%%%%%%%%%%%%%%%%%%%%%%%%%%%%%%%%%%%%%%%%%%

We classify time series from four real-world datasets: 
\begin{inparaenum}[(i)]
  \item \texttt{PenDigits}~\citep{Dua2019},
  \item \texttt{CharacterTrajectories}~\citep{Dua2019},
  \item \texttt{LSST}~\citep{allam2018photometric}, and
  \item \texttt{Physionet2012}~\citep{goldberger2000physiobank}.
\end{inparaenum}
For dataset statistics and necessary filtering steps, please refer to Section \ref{supp: Dataset stats} in the appendix.
Moreover, to efficiently compute the signature, we
sample the imputed path in a \emph{fixed} time resolution\footnote{For Physionet2012 hourly, for the other datasets once per originally observed time step}, resulting in
a piecewise linear path.
For time series that are not irregularly spaced~(this applies to all datasets but \texttt{Physionet2012}), we employ two types of random subsampling as an additional
preprocessing step,
namely
\begin{inparaenum}[(1)]
    \item `Random': Missing at random; on the instance level, we discard 50\% of all observations.
    \item `Label-based': Missing not at random; for each
      class, we uniformly sample missingness frequencies between $40\%$
      and $60\%$.
\end{inparaenum}
Since \texttt{PenDigits} consists of particularly short time series~(8
steps, 2 dimensions), we use more moderate frequencies of $30\%$
and $20$--$40\%$, respectively, for discarding observations.
Finally, we standardise all time series channels using the empirical mean and standard deviation as determined on the entire training split.

\paragraph{Models}
We study the following models:
\begin{inparaenum}[(1)]
  \item \textsc{Sig}, a simple signature model that involves a linear
    augmentation, the signature transform~(signature block) and a final
    module of dense layers,
  \item \textsc{RNN}, an RNN model using GRU cells~\citep{cho2014learning},
  \item \textsc{RNNSig}, which extends the signature transform to a window-based
    stream of signatures, and where the final neural module is a GRU
    sliding over the stream of signatures, and
  \item \textsc{DeepSig}, a deep signature model sequentially employing two signature blocks featuring augmentation and signature transforms,
  following \citet{kidger2019deep}.
\end{inparaenum}
Please refer to Supplementary Section~\ref{supp: Model Architectures} for more details about the
architectures and implementations. We use the `Signatory' package to
calculate the signature transform~\citep{signatory}, and implemented all GP
adapters using the `GPyTorch' framework~\citep{gardner2018gpytorch}.

\paragraph{Training and evaluation}
We use the predefined training and testing splits for each dataset,
separating $20\%$ of the training split as a validation set for
hyperparameter tuning.
For each setting, we run a randomised hyperparameter search of $20$
calls and train each of these fits until convergence~(at most 100
epochs; we stop early if the performance on the validation
split does not improve for 20 epochs). As for performance metrics, for
binary classification tasks, we optimise area under the precision-recall curve (as approximated via average precision) and also report AUROC. For multi-class
classification, we optimise balanced accuracy~(BAC) and additionally
report accuracy and weighted AUROC~(w-AUROC)\footnote{%
  AUROC is computed for each label and averaged with weights according to
  the support of each class%
}.
Having selected the best hyperparameter configuration for each setting,
we repeat $5$ fits; for each fit, we select the best model state in
terms of the best validation performance, and finally report mean and
standard deviation~(error bars) of the performance metrics on the
testing split.

\begin{table}[tbp] %
    \caption{\texttt{CharacterTrajectories} dataset under label-based subsampling. The top three methods are highlighted: bold \& underlined, bold, underlined. All measures are reported as percentage points. Balanced accuracy (BAC) is the metric we optimised for. We further report accuracy and weighted AUROC (w-AUROC).}
    \centering
    \input{tables/repetitions_CharacterTrajectories_Label-based}
    \label{tab:character}
\end{table}

\begin{table}[tbp]
    \caption{\texttt{PenDigits} dataset under label-based subsampling. The top three methods are highlighted: bold \& underlined, bold, underlined. All measures are reported as percentage points. Balanced accuracy (BAC) is the metric we optimised for. We further report accuracy and weighted AUROC (w-AUROC)}
    \centering
    \input{tables/repetitions_PenDigits_Label-based}
    \label{tab:pendigits}
\end{table}

\paragraph{Results}
\begin{figure*}[tbp]
  \subcaptionbox{CharacterTrajectories-R}{%
    \includegraphics[width=0.48\linewidth]{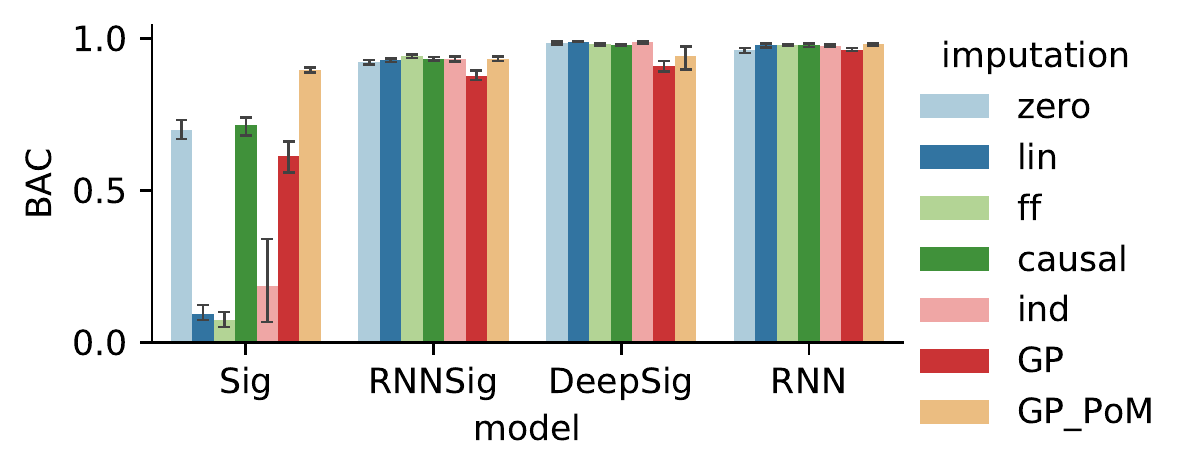}\quad%
   }%
  \subcaptionbox{CharacterTrajectories-L}{%
    \includegraphics[width=0.48\linewidth]{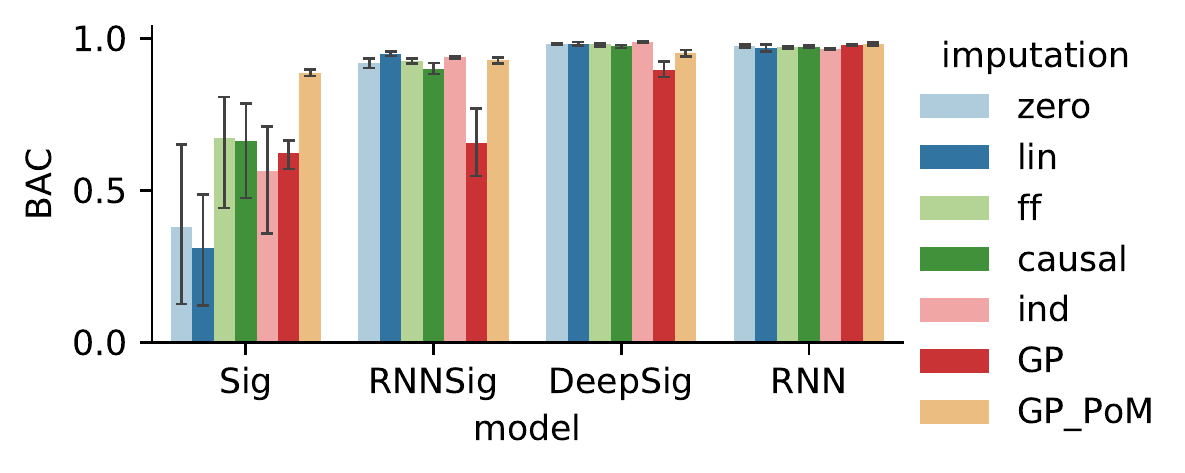}\quad%
  }
  \caption{Experimental results visualised for \texttt{CharacterTrajectories} dataset.
  The bars display performance in terms of balanced accuracy (BAC),
whereas the panels indicate the subsampling strategy.
  Left: Random subsampling (R), right: label-based subsampling (L).}
  \label{fig:results_main}
\end{figure*}

In Tables \ref{tab:character} and \ref{tab:pendigits}, the results for
\texttt{CharacterTrajectories} and \texttt{PenDigits} under label-based
subsampling are shown, respectively. For the remaining datasets and
subsampling strategies, please refer to Tables~\ref{supp: tab
character}--\ref{supp: tab physionet} in the appendix. 
We observe that both \textsc{DeepSig} as well as the signature-free \textsc{RNN} perform well
over many scenarios. In particular, they are impervious to the choice of
several imputation schemes in the sense that it does not have a large
impact on their predictive performance.
However, we also see that certain signature models, in particular
\textsc{Sig}, are heavily impacted by the choice of imputation strategy.
Figure~\ref{fig:results_main} exemplifies this finding in a barplot
visualization; for the remaining visualizations, including the number of
parameters of the optimised models, please refer to Supplementary
Figures~\ref{supp: barplots1} and~\ref{supp: barplots2}. In the case of
\texttt{CharacterTrajectories}, \textsc{Sig} was only able to achieve
acceptable performance through our novel \textsc{GP-PoM} strategy. In
\texttt{PenDigits}, we encountered issues of numerical stability for the
original GP adapter\footnote{They were addressed by jittering the
diagonal in the Cholesky decomposition.}; not so for \textsc{GP-PoM}.
Furthermore, we found that \textsc{GP-PoM} tends to converge faster to
a better performance than the original GP adapter, as exemplified in
Supplementary Figure~\ref{supp: gp-training}.

%%%%%%%%%%%%%%%%%%%%%%%%%%%%%%%%%%%%%%%%%%%%%%%%%%%%%%%%%%%%%%%%%%%%%%%%
\section{Discussion}
%%%%%%%%%%%%%%%%%%%%%%%%%%%%%%%%%%%%%%%%%%%%%%%%%%%%%%%%%%%%%%%%%%%%%%%%

Our findings suggest that the choice of path imputation strategy can
\emph{drastically} affect the performance of signature-based models. We
observe this most prominently in `shallow' signature models, whereas deep signature models~(\textsc{DeepSig})
are more robust in tackling irregular time series over different
imputations---comparable to non-signature RNNs, yet on average being
more parameter-efficient.

Overall, we find that uncertainty-aware approaches~(indicator
imputation and \textsc{GP-PoM}) are beneficial when imputing
irregularly-spaced time series for classification.
Crucially, uncertainty information has to be accessible during the
\emph{prediction step}. We find that this is indeed not the case for
the standard GP adapter~(despite the naming of `uncertainty-aware
framework'), since for each MC sample, the downstream classifier has no access
to missingness or uncertainty about the underlying imputation.
\textsc{GP-PoM}, our proposed end-to-end imputation strategy, shows
competitive classification performance, while considerably improving
upon the existing GP adapter. As for
limitations, \textsc{GP-PoM} sacrifices the GP adapter's ability to be
explicitly uncertain \emph{about} its own prediction~(due to the
variance of the MC sampled predictions), while the subsequent classifier
has to be able to handle the doubled feature dimensionality.

\paragraph{Recommendations for the practitioner}
When dealing with a challenging time series classification task, we recommend to consider signatures as a powerful tool to encode paths with little loss of information. However, we observe that this comes at a certain cost: since the signature describes continuous paths~(and not discrete time series), constructing this path from raw data is a delicate task that can heavily impact the signature and the performance of downstream models. To this end, we recommend using \textsc{GP-PoM}, which explicitly captures uncertainty in the imputed path. Given our findings, indicator imputation is a simple but promising go-to strategy, however we caution its use together with shallow signature models since we observed detrimental effects in terms of predictive performance.
Furthermore, when applying signatures in online applications or settings, where during training no data should leak from the future~(e.g.\ in online settings, this could impair performance upon deployment), we recommend to use causal~(or time-joined) path imputations: their design specifically prevents leakage from the future, even if the signature interprets the imputations as knots of a piece-wise linear path. 

\section{Conclusion}

The signature transform has recently gained attention for being
a promising feature extractor that can be easily integrated to neural
networks. As we empirically demonstrated in this paper, the application
of signature transforms to real-world temporal data is fraught with
pitfalls---specifically, we found the choice of an imputation scheme to
be crucial for obtaining high predictive performance. Moreover, by
integrating uncertainty to the prediction step, our proposed
\textsc{GP-PoM} has demonstrated overall competitive performance and in
particular improved robustness in signature models when classifying irregularly-spaced and asynchronous time series. 

\section*{Broader Impact}
Whilst the task of converting observed data into a path in data space is particularly important for signatures, it also arises in the context of, for example, convolutional and recurrent neural networks.

Convolutions are often thought of in terms of discrete sums, but they are perhaps more naturally described as the integral cross-correlation between the underlying data path $f$ and the learnt filter $g_\theta$. Given sample points $t_1, \ldots, t_n \in [0, T]$, this integral is then approximated via numerical quadrature:
\begin{equation*}
    \frac{1}{T}\int_0^T f(t) g_\theta(t) \mathrm{d}t \approx \frac{1}{n}\sum_{i = 1}^n f(t_i) g_\theta(t_i),
\end{equation*}
although the $1/n$ scaling is really only justified in the case that the $t_i$ are equally spaced.\footnote{The $g_\theta$ is typically a step function in `normal' convolutional layers. Some works exists on replacing it with e.g. B-splines \cite{fey2018splinecnn} to better handle irregular data. The oddity of scaling by $1/n$ with irregular data has not been explicitly addressed in the literature, at least to our knowledge; indeed quite conversely we have seen it used without remark.} Thus we see that with convolutions, we are implicitly interpreting the observed data as a path in data space.

Similarly, the connections between dynamical systems and recurrent neural
networks are well known~\citep{FUNAHASHI1993801, continuousrnn}, and these tend to use a similar setup.
For non-signature methods as for signature methods, this implicit usage of data as a path in data space often seems to be swept under the rug, and we have demonstrated that this is deserving further attention.

With respect to ethical considerations, we acknowledge that time series models in general can be used for better or for worse. On top of that, implicit biases underlying models as well as datasets can have unintended harmful consequences. By shedding light on the impact of implicit usage of paths in data space we hope to forward understanding and accountability of black-box models. Even if our work focuses on signature models, we believe that the principle of making underlying assumptions explicit is relevant far beyond this class of models. 

\section{Acknowledgements}
M.M, M.H, and C.B were supported by the SNSF Starting Grant `Significant
Pattern Mining'. The authors are grateful to Patrick Kidger for valuable discussions, guidance, and code contributions.

\bibliography{ref}
\bibliographystyle{apalike}
\newpage
\appendix

\section{Appendix}\label{sec:Appendix}

\subsection{Further Experiments}\label{supp:Experiments}

\begin{table}[htbp]
    \begin{center}
	\caption{\texttt{CharacterTrajectories}, random subsampling}
	\input{tables/repetitions_CharacterTrajectories_Random}
	\label{supp: tab character}
	\end{center}
\end{table}

\begin{table}[htbp]
    \begin{center}
	\caption{\texttt{PenDigits}, random subsampling}
	\input{tables/repetitions_PenDigits_Random}
	\label{supp: tab pendigits}
	\end{center}
\end{table}

\begin{table}[htbp]
    \begin{center}
	\caption{\texttt{LSST}, label-based subsampling}
	\input{tables/repetitions_LSST_Label-based}
	\label{supp: tab lsst label-based}
	\end{center}
\end{table}

\begin{table}[htbp]
    \begin{center}
	\caption{\texttt{LSST}, random subsampling}
	\input{tables/repetitions_LSST_Random}
	\label{supp: tab lsst random}
	\end{center}
\end{table}

\begin{table}[htbp]
    \begin{center}
	\caption{\texttt{Physionet 2012} }
	%\small{
	\input{tables/repetitions_Physionet2012}
	\label{supp: tab physionet}
	%}
	\end{center}
\end{table}

\begin{figure}[tbp]
\subcaptionbox{CharacterTrajectories-R}{%
    \includegraphics[width=0.48\linewidth]{plots/barplot_main_CharacterTrajectories-MissingAtRandomSubsampler.pdf}\quad%
     \includegraphics[width=0.48\linewidth]{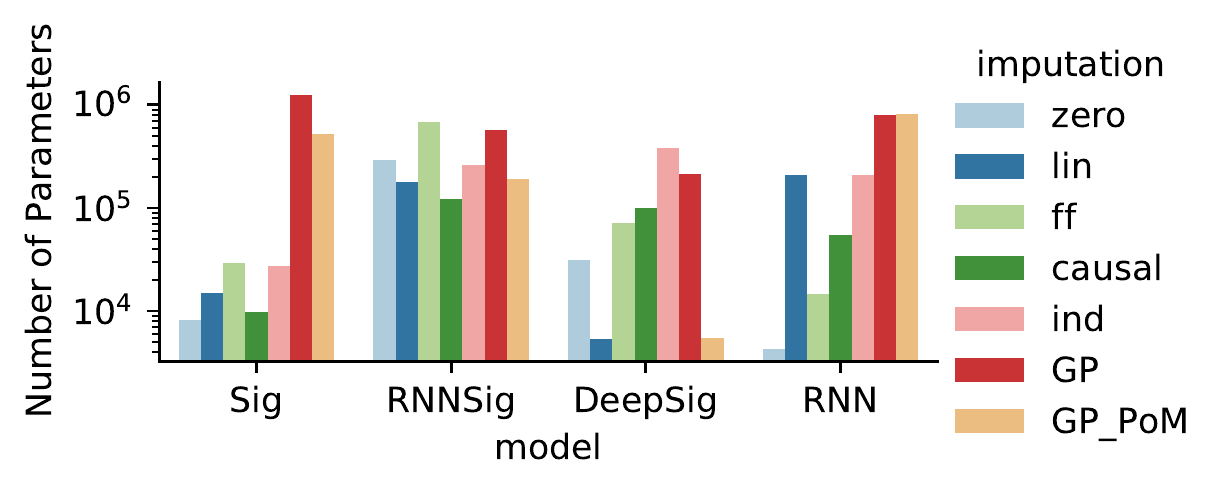}%
   }\\%
  \subcaptionbox{CharacterTrajectories-L}{%
    \includegraphics[width=0.48\linewidth]{plots/barplot_main_CharacterTrajectories-LabelBasedSubsampler.pdf}\quad%
     \includegraphics[width=0.48\linewidth]{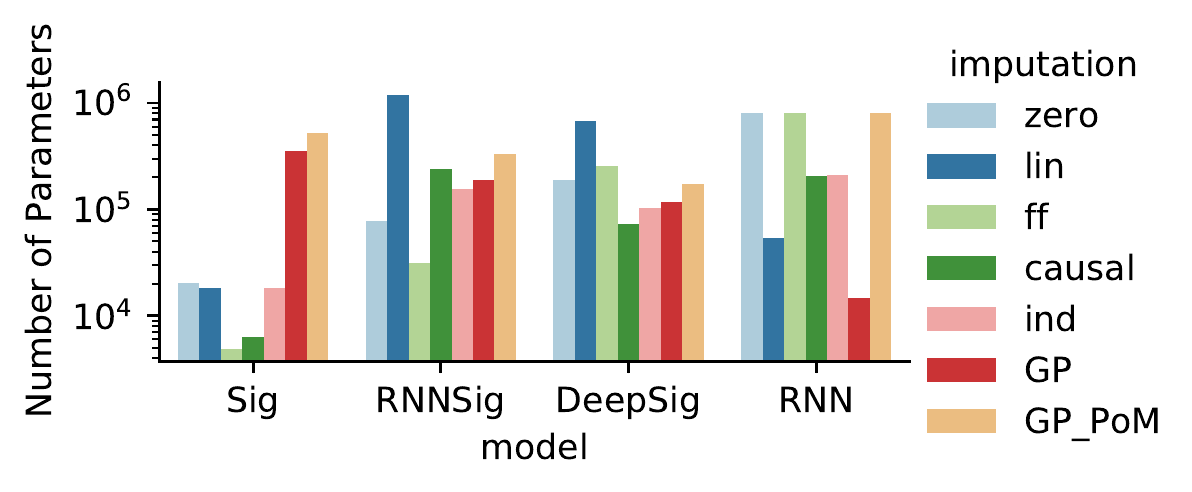}%
  }
  \subcaptionbox{LSST-R}{%
    \includegraphics[width=0.48\linewidth]{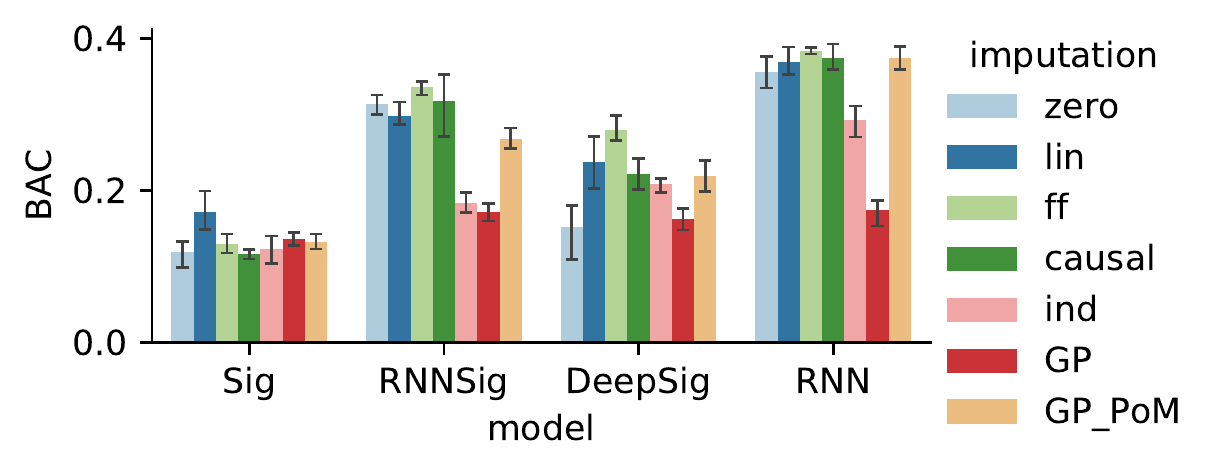}\quad%
    \includegraphics[width=0.48\linewidth]{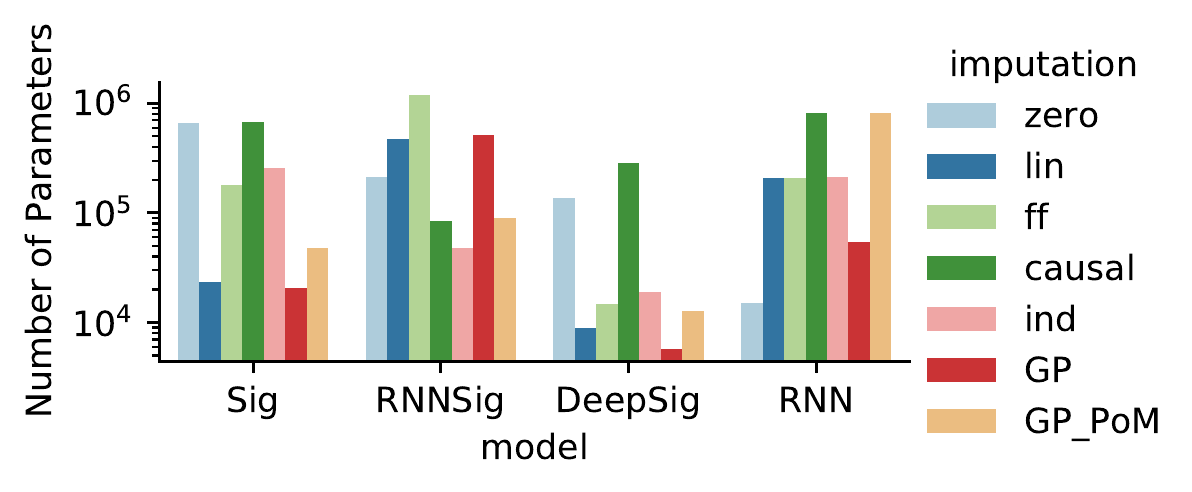}%
  }\\%
  \subcaptionbox{LSST-L}{%
    \includegraphics[width=0.48\linewidth]{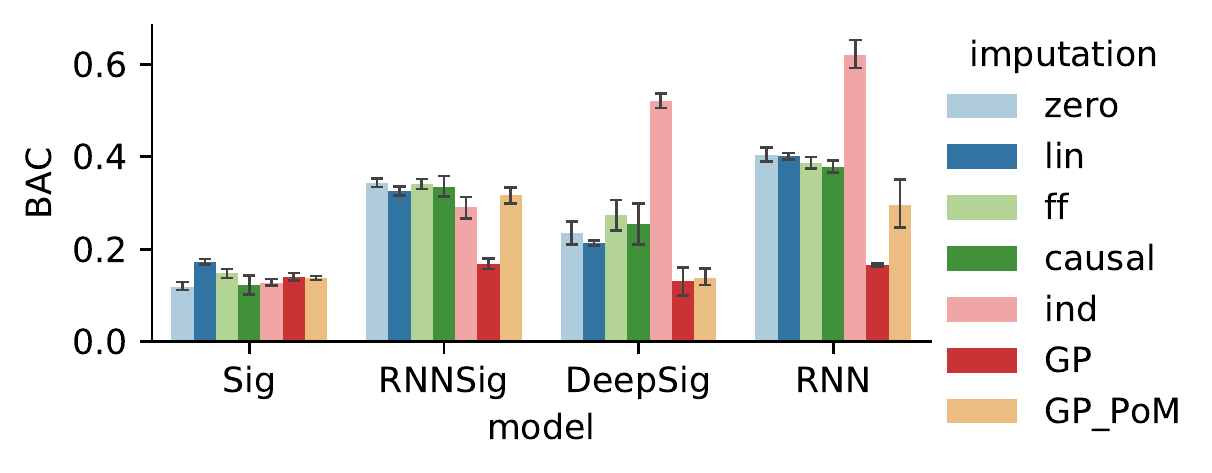}\quad%
    \includegraphics[width=0.48\linewidth]{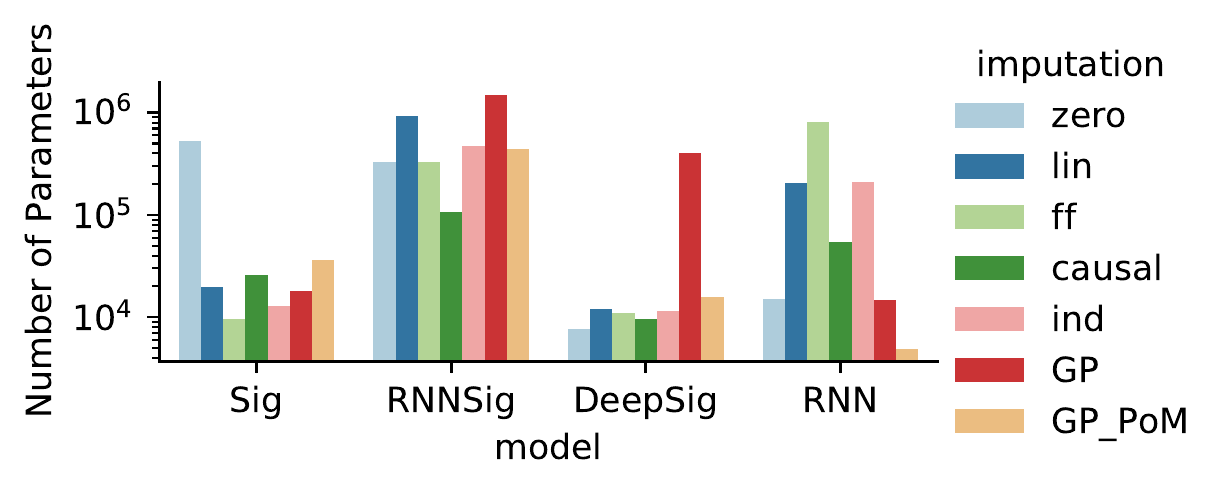}
  }%
  \caption{%
    Visualisations for \texttt{CharacterTrajectories} and \texttt{LSST} . The rows indicate datasets and different subsampling schemes (R for Random, L for Label-based). The left column displays the performance metric which was optimzied for: balanced accuracy (BAC), or average precision. The right column indicates the number of trainable parameters which the best model required (as selected in the hyperparameter search).
  }
  \label{supp: barplots1}
\end{figure}
\begin{figure}
   \subcaptionbox{PenDigits-R}{%
     \includegraphics[width=0.48\linewidth]{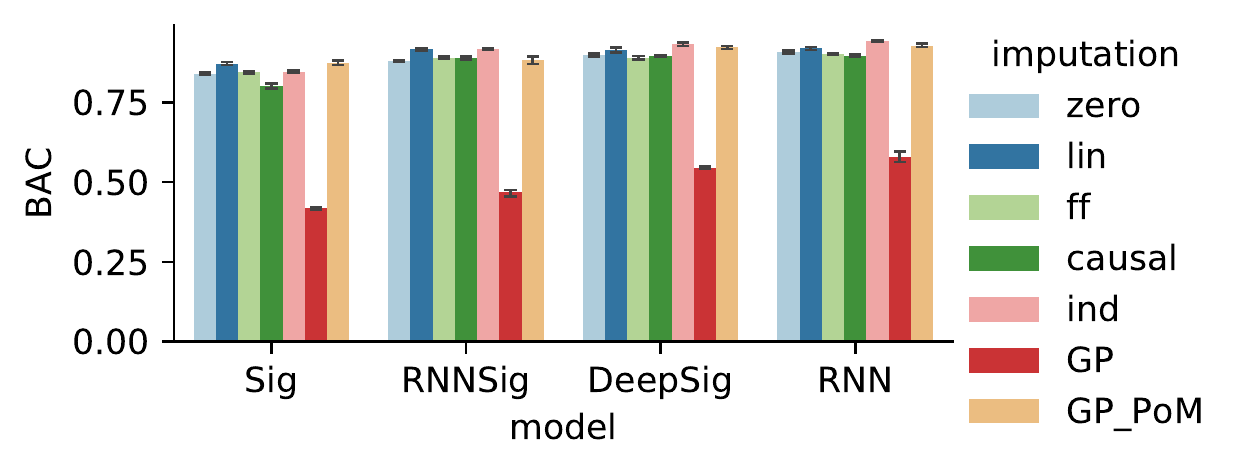}\quad%
     \includegraphics[width=0.48\linewidth]{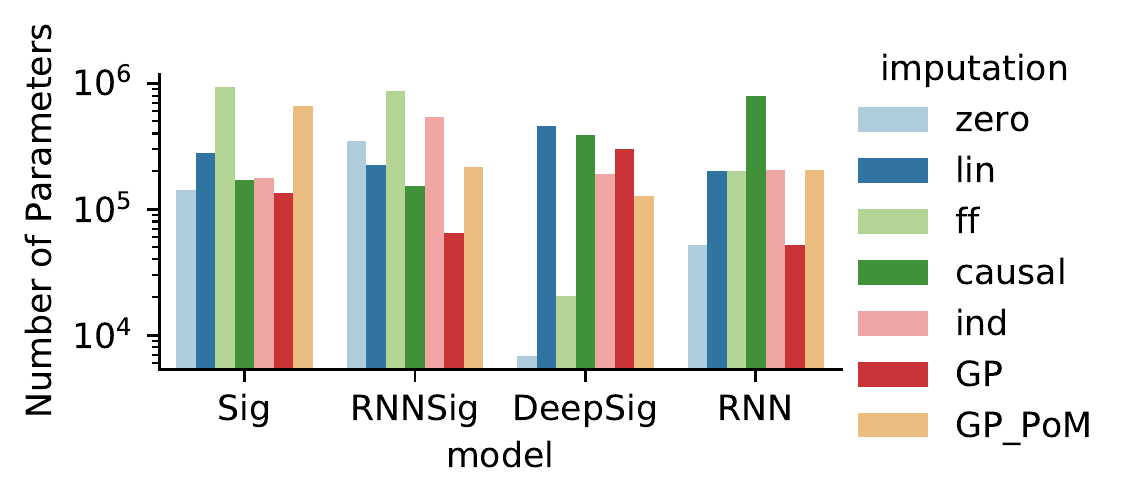}
   }\\%
  \subcaptionbox{PenDigits-L}{%
    \includegraphics[width=0.48\linewidth]{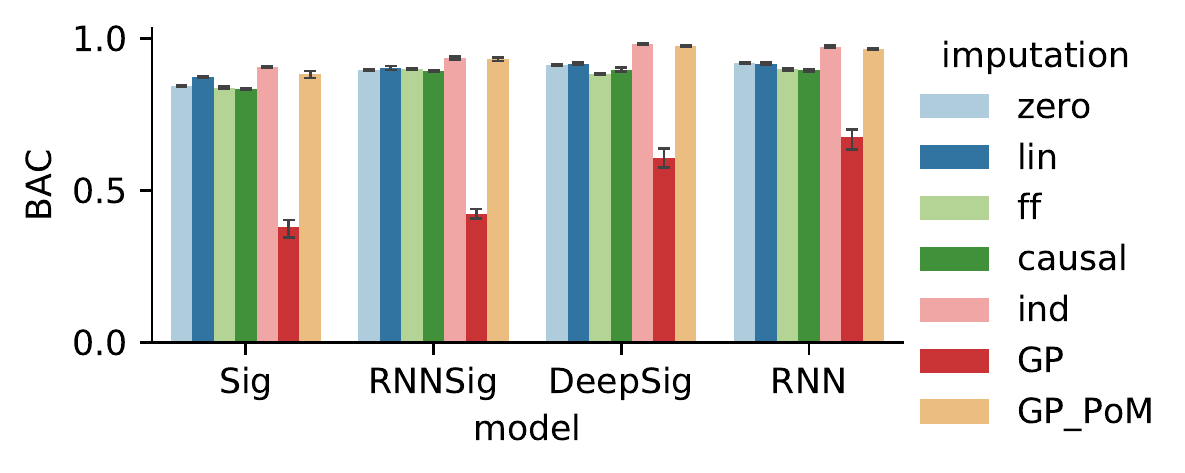}\quad%
    \includegraphics[width=0.48\linewidth]{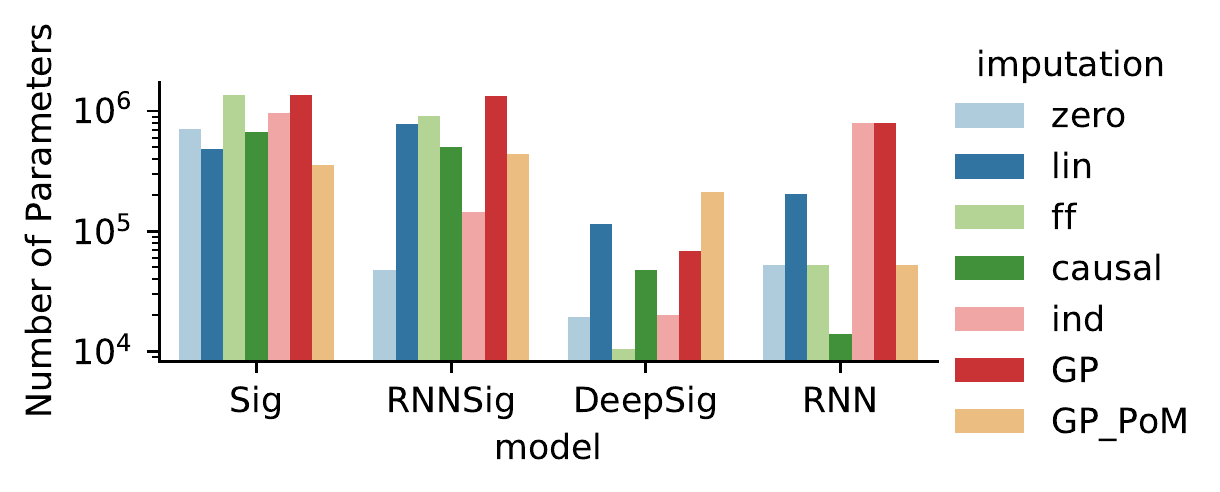}
  }\\ %
  \subcaptionbox{Physionet2012}{%
    \includegraphics[width=0.48\linewidth]{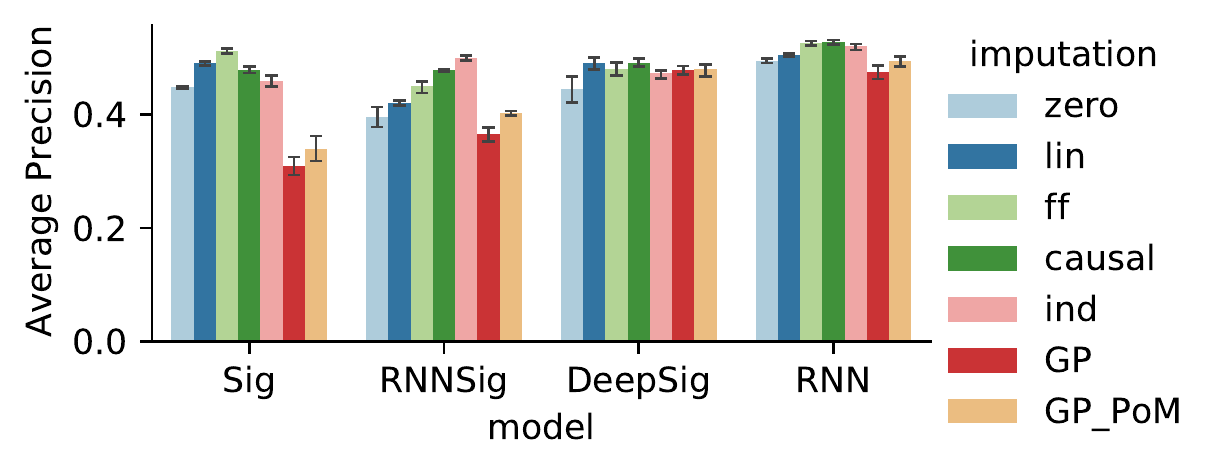}\quad%
    \includegraphics[width=0.48\linewidth]{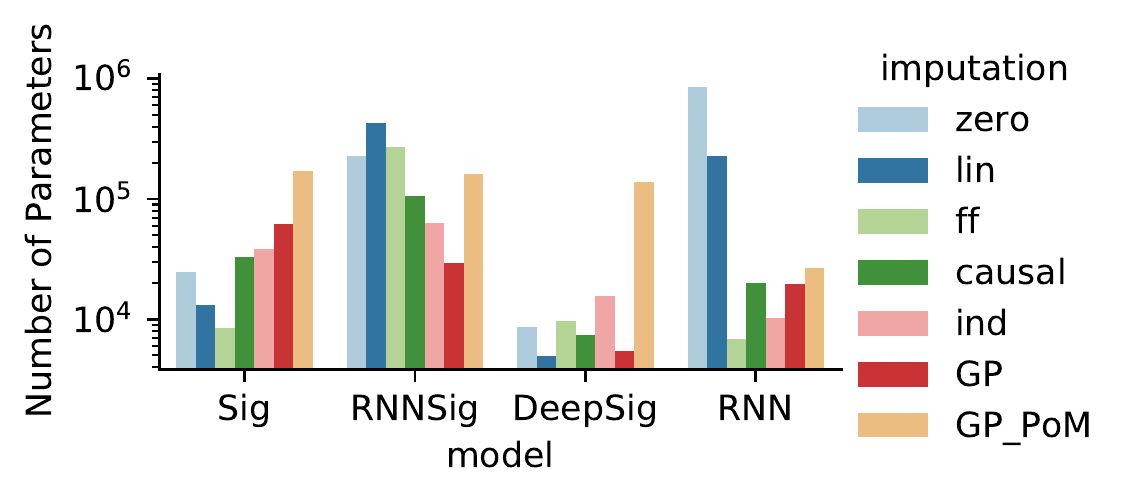}
  }
  \caption{%
    Visualisations for \texttt{PenDigits} and \texttt{Physionet} . The rows indicate datasets and different subsampling schemes (R for Random, L for Label-based). The left column displays the performance metric which was optimzied for: balanced accuracy (BAC), or average precision. The right column indicates the number of trainable parameters which the best model required (as selected in the hyperparameter search).
  }
  \label{supp: barplots2}
\end{figure}

\begin{figure}[tbp] 
    \begin{center}
    \includegraphics[width=0.95\linewidth]{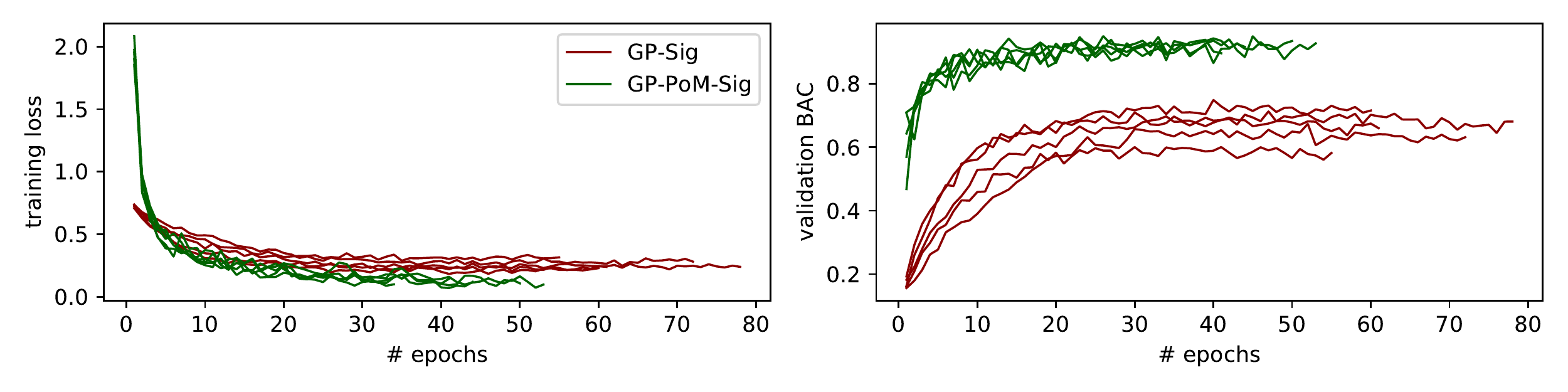}\quad%
  \end{center}
  \caption{GP-PoM training illustrated for CharacterTrajectories as compared to conventional GP adapter.}
  \label{supp: gp-training}
\end{figure}

%%%%%%%%%%%%%%%%%%%%%%%%%%%%%%%%%%%%%%%%%%%%%%%%%%%%%%%%%%%%%%%%%%%%%%%%
\subsection{Imputation strategies} \label{supp: Imputation}
%%%%%%%%%%%%%%%%%%%%%%%%%%%%%%%%%%%%%%%%%%%%%%%%%%%%%%%%%%%%%%%%%%%%%%%%

We consider the following set of strategies for path imputation, i.e.\
\begin{enumerate}
    \item linear interpolation: At a given imputation point, the previous and next observed data point are linearly interpolated. Missing values at the start or end of the time series are imputed with $0$ which for standardised data also corresponds to the mean.
    \item forward filling: At a given imputation point, the last observed value is carried forward. Missing values at the start of the time series are imputed with $0$.
    \item indicator imputation: At a given imputation point, for each feature dimension, if no observation is available a binary missingness indicator variable is set to $1$, $0$ otherwise. The missing value is filled with $0$.
    \item zero imputation: At a given imputation point, missing values are filled with $0$.
    \item causal imputation: This approach is related to forward filling
      and motivated by signature theory. As opposed to forward filling,
      the time and the actual value are updated sequentially. For more
      details, we introduce causal imputation in Section~\ref{sec:Causal signature imputation}.
    \item Gaussian process adapter: We introduce GP adapters in
      Section~\ref{sec: GPadapter}, where $\mathbf{z}$
      refers to the imputed time series~(modelled as Gaussian
      distribution).
\end{enumerate}

%%%%%%%%%%%%%%%%%%%%%%%%%%%%%%%%%%%%%%%%%%%%%%%%%%%%%%%%%%%%%%%%%%%%%%%%
\subsection{Dataset statistics and filtering} \label{supp: Dataset stats}
%%%%%%%%%%%%%%%%%%%%%%%%%%%%%%%%%%%%%%%%%%%%%%%%%%%%%%%%%%%%%%%%%%%%%%%%
\paragraph{Physionet2012}
As our focus is time series classification, for \texttt{Physionet2012}~\citep{goldberger2000physiobank}, we included the 36 time series variables, and excluded the static covariates (notably, we counted the variable `weight' as a static covariate). Subsequently, we excluded the following $12$ icu stays (here represented by there ids) for having no time series data (but only static covariates): $140501, 150649, 140936, 143656, 141264, 145611, 142998, 147514,$  $142731, 150309, 155655, 156254 $, and a single noisy encounter, $135365$, which contained much more observations than all other patients. After these filtering steps, we count $11987$ instances and a binary class label, whether a patient survives the hospital stay or not.

\paragraph{PenDigits}
For \texttt{PenDigits}~\citep{Dua2019}, we count $10992$ samples, featuring $2$ channels and $8$ time steps, and $10$ classes.

\paragraph{LSST}
LSST~\citep{allam2018photometric} contains $4925$ instances featuring $6$ channel dimensions and $36$ time steps. This dataset contains $14$ classes.

 \paragraph{CharacterTrajectories}
 This dataset contains $2858$ instances, featuring $3$ channel dimensions, $182$ time steps and $20$ classes~\citep{Dua2019}.

%%%%%%%%%%%%%%%%%%%%%%%%%%%%%%%%%%%%%%%%%%%%%%%%%%%%%%%%%%%%%%%%%%%%%%%%
\subsection{Model implementations, architectures and hyperparameters} \label{supp: Model Architectures}
%%%%%%%%%%%%%%%%%%%%%%%%%%%%%%%%%%%%%%%%%%%%%%%%%%%%%%%%%%%%%%%%%%%%%%%%

All models are implemented in Pytorch~\citep{pytorch2019}, whereas the
GP adapter and \textsc{GP-PoM} are implemented using the GPyTorch
framework~\citep{gardner2018gpytorch}. Next, we specify the details of
the model architectures.

\paragraph{\textsc{Sig}}
We use a simple signature model that involves one        signature block comprising of a linear
    augmentation followed by the signature transform. Subsequently, a final module of dense layers $(30,30)$ is used. This is architecture refers to the Neural-signature-augment model \cite{kidger2019deep}.
  \paragraph{\textsc{RNNSig}} This model extends the signature transform to a window-based
    stream of signatures, where the final neural module is a GRU
    sliding over the stream of signatures. We allowed window sizes between $3$ and $10$ steps. For the GRU cell, we allowed any of the following number of hidden units: $[16,32,64,128]$.
    
  \paragraph{\textsc{RNN}} Here, we use a standard RNN model using GRU cells. The size of hidden units was chosen as one of the following: $[16,32,64,128, 256, 512]$.
  \paragraph{\textsc{DeepSig}} For the deep signature model we  employ two signature blocks (each comprising a linear augmentation and the signature calculation) following \citet{kidger2019deep}.

\subsubsection{Hyperparameters}

For all signature-based models, we allowed a signature truncation
depth of $2$--$4$, as we observed that larger values quickly led to
a parameter explosion. All models were optimised using Adam
\citep{kingma2014adam}. Both the learning rates and weight decay
were drawn log-uniformly between $10^{-4}$ and $10^{-2}$. We allowed
for the following batch-sizes: $(32, 64, 128, 256)$. For GP-based
models, to save memory, we used virtual batching based on
a batch-size of~$32$. Furthermore, for standard GP adapters we used $10$ MC samples, conforming with recent literature \citep{futoma2017mgp, moor2019early}. All approaches were constrained to have no more than $1.5$ million trainable parameters.

%%%%%%%%%%%%%%%%%%%%%%%%%%%%%%%%%%%%%%%%%%%%%%%%%%%%%%%%%%%%%%%%%%%%%%%%
\subsection{Fragile dependence on sampling in unrelated channels: example}
\label{sec:Fragile dependence} 
%%%%%%%%%%%%%%%%%%%%%%%%%%%%%%%%%%%%%%%%%%%%%%%%%%%%%%%%%%%%%%%%%%%%%%%%

%%%%%%%%%%%%%%%%%%%%%%%%%%%%%%%%%%%%%%%%%%%%%%%%%%%%%%%%%%%%%%%%%%%%%%%%
\begin{figure}[t]
    \centering
    \includegraphics[width=0.35 \columnwidth]{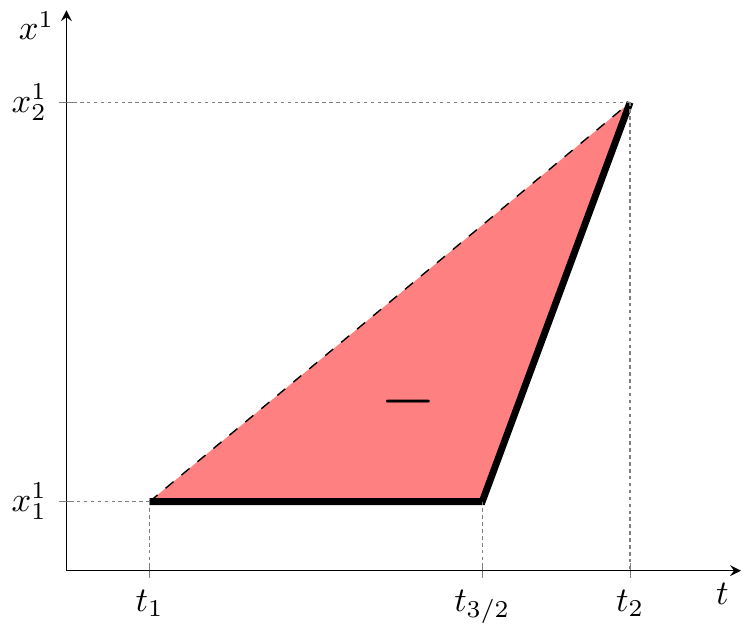}
    \caption{
      L{\'e}vy area of the forward-fill imputed path. By changing
      $t_{3/2}$~(a \emph{single} unrelated observation!), we can make
      this disparity greater or smaller.
    }
    \label{fig:bentline}
\end{figure}
%%%%%%%%%%%%%%%%%%%%%%%%%%%%%%%%%%%%%%%%%%%%%%%%%%%%%%%%%%%%%%%%%%%%%%%%

Suppose that we have observed the~(very short) time series
\begin{equation}\label{eq:flaw1}
    \mathbf{x} = ((t_1, x_1^1, x_1^2), (t_2, x_2^1, *)) \in \seriesspace(\reals^2).
\end{equation}
Perhaps we now apply, say, forward fill data-imputation, to produce
\begin{equation*}
    ((t_1, x_1^1, x_1^2), (t_2, x_2^1, x_1^2)).
\end{equation*}
Finally we linearly path-impute to create the linear path
\begin{align*}
    f &\colon [t_1, t_2] \to \reals \times \reals^2\\
    f &\colon t \mapsto \left(t, x_1^1\frac{t_2 - t}{t_2 - t_1} + x_2^1\frac{t - t_1}{t_2 - t_1}, x_1^2\right),
\end{align*}
to which we may then apply the signature transform. In particular we
will have computed the L{\'e}vy area with respect to $t$ and $x^1$. As this
is just a straight line, the L{\'e}vy area is zero.

Now suppose we include an additional observation at some time $t_{3/2} \in (t_1, t_2)$, so that our data is instead
\begin{equation}\label{eq:flaw2}
    \mathbf{x} = ((t_1, x_1^1, x_1^2), (t_{3/2}, *, x_{3/2}^2), (t_2, x_2^1, *)).
\end{equation}
Then the same procedure as before will produce the data
\begin{equation*}
    \mathbf{x} = ((t_1, x_1^1, x_1^2), (t_{3/2}, x_1^1, x_{3/2}^2), (t_2, x_2^1, x_{3/2}^2)),
\end{equation*}
with corresponding function $f$. The $(t, x^1)$ components of $f$ and
its $(t, x^1)$-L{\'e}vy area are shown in Figure~\ref{fig:bentline}. As
a result of an unrelated observation in the $x^2$ channel, the $(t,
x^1)$-L{\'e}vy area has been changed.
The closer $t_{3/2}$ is to $t_2$, the greater the disparity.
This simple example underscores the danger of `just forward-fill
data-imputing'. Doing so has introduced an undesired dependency on the
simple \emph{presence} of an observation in other channels, with the
change in our imputed path being determined by the \emph{time} at which
this other observation occurred.

Indeed, \emph{any} imputation scheme that predicts something other than
the unique value lying on the dashed line in Figure~\ref{fig:bentline},
will fail. This means that this example holds for essentially every
data-imputation scheme---the only scheme that survives this flaw is the
linear data-imputation scheme. This is the unique imputation scheme
that coincides with the linear path-imputation that \emph{must} be our
concluding step.
However, when there is missing data at the start or the end of
a partially observed times series, then there is no `next observation'
which linear imputation may use. So in general, we cannot uniformly
apply the linear data-imputation scheme, and must choose another scheme or find ad-hoc solutions for missing data at the start or the end of the time series. Furthermore, it is plausible to assume that linear interpolation suffers from low expressivity as an imputation scheme which might empirically mask this benefit.

%%%%%%%%%%%%%%%%%%%%%%%%%%%%%%%%%%%%%%%%%%%%%%%%%%%%%%%%%%%%%%%%%%%%%%%%
\subsection{Causal signature imputation}\label{sec:Causal signature imputation}
%%%%%%%%%%%%%%%%%%%%%%%%%%%%%%%%%%%%%%%%%%%%%%%%%%%%%%%%%%%%%%%%%%%%%%%%

In Section \ref{sec:Fragile dependence} we have spoken about the limitations of traditional data-imputation schemes, and at first glance one may be forgiven for
thinking that these are issues are unavoidable. 
However, it turns out that we need not be limited just to these
traditional imputation schemes. The trick is to consider time not as
a \emph{parameterisation}, but as a \emph{channel}\footnote{To be clear, using time as a channel is already
a well-known trick in the signature literature that we do not take
credit for inventing! See for example \citet[Definition
A.3]{kidger2019deep}. It is however pleasing that something commonly
used in the theory of signatures is also what allows us to overcome what
we identify as some of their limitations.}.
This leads to a `meta imputation strategy', which we refer to as
\emph{causal signature imputation}. It will turn any traditional causal
data-imputation strategy (for example, feed-forward) into a causal
path-imputation strategy for signatures; at the same time it will
overcome the issue of a fragile dependence.

Suppose we have $\mathbf{x} \in \seriesspace(\dataspace^*)$, and some
favourite choice of causal data-imputation strategy $c \colon
\seriesspace(\dataspace^*) \to \seriesspace(\dataspace)$.
Next, given
\begin{align}
    \mathbf{x} = ((t_1, x_1), \ldots, (t_n, x_n)) \in \seriesspace(\dataspace),
\end{align}
we define the operation $\Omega \colon \seriesspace(\dataspace) \to \seriesspace(\dataspace)$ by
\begin{align}
    \Omega(\mathbf{x}) = (&(t_1, x_1), (t_2, x_1), (t_2, x_2),(t_3, x_2),\nonumber\\
    &\ldots,\nonumber\\
    &(t_i, x_i), (t_{i + 1}, x_i), (t_{i + 1}, x_{i + 1}), (t_{i + 2}, x_{i + 1}),\nonumber\\
    &\ldots,\nonumber\\
    &(t_{n - 1}, x_{n - 1}), (t_n, x_{n - 1}),(t_n, x_n)).\label{eq:causalsig}
\end{align}
That is, \emph{first} time is updated, and \emph{then} the corresponding observation
in data space is updated. This means that the change in data space
occurs instantaneously.

For each $n \in \naturals$~(and given $a < b$), fix any $s_i^{(n)}$ for
$i \in \{1, \ldots, n \}$. 
(We will see that the exact choice is unimportant in a moment.)
Given
\begin{align*}
    \mathbf{x} = ((t_1, x_1), \ldots, (t_n, x_n)) \in \seriesspace(\dataspace),
\end{align*}
let $\psi \colon \seriesspace(\dataspace) \to (\reals \times
\dataspace)^{[a, b]}$ be the unique continuous piecewise linear path
such that $\psi(s_i^{(n)}) = (t_i, x_i)$. Note that this is just
a slight generalisation of the linear path-imputation that has already
been performed so far; we are simply no longer asking for additional
assumptions of the form $s_i^{(n)} = t_i$.\footnote{As in the
$\mathbf{\varphi}_\theta$ of \cite{toth2019gp}, for example.}

Finally, we put this all together, and define the causal signature imputation strategy $\phi_c$ associated with $c$ to be
\begin{equation*}
\phi_c = \psi \circ \Omega \circ c,
\end{equation*}
which will be a map $\seriesspace(\dataspace^*) \to (\reals \times \dataspace)^{[a, b]}$.
Thus $\phi_c$ defines a family of path-imputation schemes, parameterised by a choice of data-imputation scheme.

Before we analyse \emph{why} this works in practice, we repeat a crucial
property of the signature transform~\citep[Appendix~A]{kidger2019deep}.
\begin{theorem}[Invariance to reparameterisation]\label{theorem:invariancetime}
  Let $f \colon [a, b] \to \reals^d$ be a continuous piecewise
  differentiable path. Let $\psi \colon [a, b] \to [c, d]$ be continuously
  differentiable, increasing, and surjective. Then $\sig(f) = \sig(f \circ
  \psi)$.
\end{theorem}
Coming back to our analysis, we first note that the previous theorem
implies that the signature transform of $\phi_c(\mathbf{x})$ is
invariant to the choice of $s_i^{(n)}$.
Second, note that holding time between observations fixed is a valid
choice, by the definition for $\seriesspace$ in equation
\eqref{eq:seriesspace}. There should hopefully be no moral objection to
our definition of $\seriesspace$, as holding time fixed essentially just
corresponds to a jump discontinuity; not such a strange thing to have
occur. Here, by replacing time as the parameterisation, we are then able
to recover the continuity of the path.
Third, we claim that  $\phi_c$ is immune to the two major flaws of
imputation methods, namely
\begin{inparaenum}[(i)]
  \item their fragile dependence on sampling in unrelated channels, and
  \item their non-causality.
\end{inparaenum}
Let us consider the first flaw of dependence on sampling in unrelated channels.
For simplicity, take $c$ to be the forward-fill data-imputation
strategy. Consider again the $\mathbf{x}$ defined in expression~\eqref{eq:flaw1}.
This means that
\begin{equation}\label{eq:causal1}
    \phi_c(\mathbf{x}) = \psi(\;((t_1, x_1^1, x_1^2), (t_2, x_1^1, x_1^2), (t_2, x_2^1, x_1^2))\;).
\end{equation}
Contrast adding in the extra observation at $t_{3/2}$ as in equation \eqref{eq:flaw2}. Then
\begin{align}
    &\phi_c(\mathbf{x})(s)\nonumber\\
    &=\psi(\;((t_1, x_1^1, x_1^2), (t_{3/2}, x_1^1, x_1^2), (t_{3/2}, x_1^1, x_{3/2}^2),\nonumber\\ &\hspace{3.1em}(t_2, x_1^1, x_{3/2}^2), (t_2, x_2^1, x_{3/2}^2))\;).\label{eq:causal2}
\end{align}
Evaluating each $\psi$ will then in each case give a path with three
channels, corresponding to $t, x^1, x^2$. Then it is clear that the $(t,
x^1)$ component of the path in equation~\eqref{eq:causal1} is just
a reparameterisation of the path in equation~\eqref{eq:causal2},
a difference which is irrelevant by Theorem~\ref{theorem:invariancetime}.
(And the $x^2$ component of the second
path has been updated to use the new information $x_{3/2}^2$.) Thus the causal path impuation scheme is robust to such issues. For general time series and
$c$ taken to be any other causal data-imputation strategy, then much the
same analysis can be easily be performed.

Now consider the second potential flaw, of non-causality. The issue
previously arose because of the non-causality of the linear
path-imputation. We see from equation \eqref{eq:causalsig}, however,
such changes only occur in data space while the time channel is frozen;
conversely the time channel only updates with the value in the data
space frozen. Provided that $c$ is also causal, then causality will,
overall, have been preserved. For example, it is possible to use this
scheme in an online setting.
There are interesting comparisons to be made between causal signature
imputation and certain operations in the signature literature. First is
the \emph{lead-lag} transform \cite{primer2016}. With the lead-lag
transform, the entire path is \emph{duplicated}, and then each side is
alternately updated. Conversely, in causal signature imputation, the
path is instead \emph{split} between $t$ and $(x^1, \ldots, x^n)$, and
then each side is alternately updated.
Second is the comparison to the linear and rectilinear embedding
strategies, see for example \cite{fermanian2019embedding}. It is
possible to interpret $\psi \circ \Phi$ as a hybrid between the linear
and rectilinear embeddings: it is rectilinear with respect to an
ordering of $t$ and $(x^1, \ldots, x^n)$, and linear on $(x^1, \ldots,
x^n)$.
Furthermore, the time-joined transformation \cite{levin2013} is pursuing a very similar goal to the here described causal signature imputation. This is also why we do not consider this imputation strategy as a novel contribution of this work.

\subsubsection{Comparison to the Fourier and wavelet transforms}\label{sec: comparison fourier}
The signature transform exhibits a certain similarity to the one-dimensional Fourier or wavelet transforms. Both are integrals of paths. However, in reality these transforms are fundamentally different. Both the Fourier and wavelet transforms are linear transforms, and operate on each channel of the input path separately. In doing so they model the path as a linear combination of elements from some basis.

Conversely, the signature transform is a nonlinear transform - indeed, it is a universal nonlinearity - and operates by combining information between different channels of the input path. In doing, the signature transform models \emph{functions of the path}; the universal nonlinearity property says that in some sense it provides a basis for such functions.

\end{document}

%% file: tables/repetitions_CharacterTrajectories_Label-based.tex
\begin{tabular}{lllll}
\toprule
Imputation     & Model   &                                        w-AUROC &                                            BAC &                                       Accuracy \\
\midrule
\multirow{4}{*}{GP-PoM}          & DeepSig &                           $ 99.582 \pm 0.671 $ &                           $ 95.155 \pm 1.501 $ &                           $ 94.958 \pm 1.716 $ \\
                                 & RNN     &            $  \underline{ 99.973 \pm 0.015 } $ &               $  \mathbf{ 98.161 \pm 0.664 } $ &               $  \mathbf{ 98.273 \pm 0.602 } $ \\
                                 & RNNSig  &                           $ 99.696 \pm 0.089 $ &                           $ 92.778 \pm 1.239 $ &                           $ 93.231 \pm 1.133 $ \\
                                 & Sig     &                           $ 99.516 \pm 0.075 $ &                           $ 88.627 \pm 1.416 $ &                           $ 89.011 \pm 1.319 $ \\
\midrule
\multirow{4}{*}{GP}              & DeepSig &                           $ 99.290 \pm 0.704 $ &                           $ 89.545 \pm 2.996 $ &                           $ 89.368 \pm 3.123 $ \\
                                 & RNN     &                           $ 99.970 \pm 0.011 $ &                           $ 97.712 \pm 0.266 $ &                           $ 97.873 \pm 0.251 $ \\
                                 & RNNSig  &                           $ 96.669 \pm 2.393 $ &                          $ 65.717 \pm 13.691 $ &                          $ 67.052 \pm 13.182 $ \\
                                 & Sig     &                           $ 95.283 \pm 1.602 $ &                           $ 62.423 \pm 6.110 $ &                           $ 63.614 \pm 5.958 $ \\
\midrule
\multirow{4}{*}{causal}          & DeepSig &                           $ 99.940 \pm 0.024 $ &                           $ 97.272 \pm 0.709 $ &                           $ 97.437 \pm 0.620 $ \\
                                 & RNN     &                           $ 99.960 \pm 0.010 $ &                           $ 97.239 \pm 0.516 $ &                           $ 97.409 \pm 0.481 $ \\
                                 & RNNSig  &                           $ 99.523 \pm 0.155 $ &                           $ 89.922 \pm 2.301 $ &                           $ 90.585 \pm 2.186 $ \\
                                 & Sig     &                           $ 95.747 \pm 4.957 $ &                          $ 66.307 \pm 21.794 $ &                          $ 68.259 \pm 20.757 $ \\
\midrule
\multirow{4}{*}{forward-filling} & DeepSig &                           $ 99.953 \pm 0.041 $ &                           $ 97.956 \pm 0.677 $ &                           $ 98.078 \pm 0.656 $ \\
                                 & RNN     &                           $ 99.942 \pm 0.011 $ &                           $ 96.942 \pm 0.486 $ &                           $ 97.159 \pm 0.444 $ \\
                                 & RNNSig  &                           $ 99.720 \pm 0.071 $ &                           $ 92.568 \pm 1.091 $ &                           $ 93.148 \pm 1.011 $ \\
                                 & Sig     &                           $ 94.828 \pm 8.117 $ &                          $ 67.169 \pm 26.338 $ &                          $ 68.649 \pm 26.125 $ \\
\midrule
\multirow{4}{*}{indicator}       & DeepSig &  $  \mathbf{ \underline{ 99.988 \pm 0.013 }} $ &  $  \mathbf{ \underline{ 98.591 \pm 0.294 }} $ &  $  \mathbf{ \underline{ 98.719 \pm 0.263 }} $ \\
                                 & RNN     &                           $ 99.916 \pm 0.020 $ &                           $ 96.414 \pm 0.406 $ &                           $ 96.671 \pm 0.367 $ \\
                                 & RNNSig  &                           $ 99.802 \pm 0.032 $ &                           $ 93.787 \pm 0.463 $ &                           $ 94.234 \pm 0.442 $ \\
                                 & Sig     &                          $ 91.661 \pm 10.003 $ &                          $ 56.423 \pm 22.796 $ &                          $ 58.384 \pm 22.932 $ \\
\midrule
\multirow{4}{*}{linear}          & DeepSig &                           $ 99.970 \pm 0.010 $ &            $  \underline{ 98.051 \pm 0.743 } $ &            $  \underline{ 98.217 \pm 0.671 } $ \\
                                 & RNN     &                           $ 99.880 \pm 0.059 $ &                           $ 96.906 \pm 1.314 $ &                           $ 97.117 \pm 1.196 $ \\
                                 & RNNSig  &                           $ 99.876 \pm 0.035 $ &                           $ 94.848 \pm 0.916 $ &                           $ 95.292 \pm 0.842 $ \\
                                 & Sig     &                          $ 80.442 \pm 18.228 $ &                          $ 31.193 \pm 23.962 $ &                          $ 32.326 \pm 24.679 $ \\
\midrule
\multirow{4}{*}{zero}            & DeepSig &               $  \mathbf{ 99.977 \pm 0.010 } $ &                           $ 98.030 \pm 0.357 $ &                           $ 98.189 \pm 0.358 $ \\
                                 & RNN     &                           $ 99.967 \pm 0.014 $ &                           $ 97.428 \pm 0.572 $ &                           $ 97.549 \pm 0.596 $ \\
                                 & RNNSig  &                           $ 99.699 \pm 0.132 $ &                           $ 91.752 \pm 1.782 $ &                           $ 92.368 \pm 1.662 $ \\
                                 & Sig     &                          $ 77.727 \pm 23.671 $ &                          $ 37.992 \pm 34.456 $ &                          $ 38.955 \pm 35.232 $ \\
\bottomrule
\end{tabular}

%% file: tables/repetitions_PenDigits_Label-based.tex
\begin{tabular}{lllll}
\toprule
Imputation                       & Model   &                                       w-AUROC &                                            BAC &                                       Accuracy \\
\midrule
\multirow{4}{*}{GP-PoM}          & DeepSig &            $  \underline{ 99.930 \pm 0.032 } $ &               $  \mathbf{ 97.403 \pm 0.300 } $ &               $  \mathbf{ 97.381 \pm 0.298 } $ \\
                                 & RNN     &                           $ 99.901 \pm 0.016 $ &                           $ 96.349 \pm 0.297 $ &                           $ 96.306 \pm 0.302 $ \\
                                 & RNNSig  &                           $ 99.669 \pm 0.073 $ &                           $ 93.022 \pm 0.765 $ &                           $ 92.967 \pm 0.763 $ \\
                                 & Sig     &                           $ 99.150 \pm 0.144 $ &                           $ 88.090 \pm 1.493 $ &                           $ 87.999 \pm 1.499 $ \\
\midrule
\multirow{4}{*}{GP}              & DeepSig &                           $ 92.885 \pm 1.455 $ &                           $ 60.593 \pm 4.092 $ &                           $ 60.476 \pm 4.067 $ \\
                                 & RNN     &                           $ 95.170 \pm 1.438 $ &                           $ 67.543 \pm 4.782 $ &                           $ 67.426 \pm 4.790 $ \\
                                 & RNNSig  &                           $ 84.501 \pm 1.307 $ &                           $ 42.184 \pm 1.977 $ &                           $ 42.141 \pm 1.913 $ \\
                                 & Sig     &                           $ 80.312 \pm 2.655 $ &                           $ 37.767 \pm 3.611 $ &                           $ 37.725 \pm 3.646 $ \\
\midrule
\multirow{4}{*}{causal}          & DeepSig &                           $ 99.241 \pm 0.075 $ &                           $ 89.616 \pm 0.749 $ &                           $ 89.514 \pm 0.747 $ \\
                                 & RNN     &                           $ 99.241 \pm 0.098 $ &                           $ 89.496 \pm 0.480 $ &                           $ 89.417 \pm 0.501 $ \\
                                 & RNNSig  &                           $ 99.298 \pm 0.041 $ &                           $ 89.187 \pm 0.476 $ &                           $ 89.137 \pm 0.494 $ \\
                                 & Sig     &                           $ 98.374 \pm 0.065 $ &                           $ 83.205 \pm 0.404 $ &                           $ 83.082 \pm 0.426 $ \\
\midrule
\multirow{4}{*}{forward-filling} & DeepSig &                           $ 99.007 \pm 0.072 $ &                           $ 88.205 \pm 0.434 $ &                           $ 88.090 \pm 0.428 $ \\
                                 & RNN     &                           $ 99.333 \pm 0.046 $ &                           $ 89.747 \pm 0.406 $ &                           $ 89.657 \pm 0.419 $ \\
                                 & RNNSig  &                           $ 99.274 \pm 0.015 $ &                           $ 89.788 \pm 0.384 $ &                           $ 89.743 \pm 0.392 $ \\
                                 & Sig     &                           $ 98.310 \pm 0.045 $ &                           $ 83.739 \pm 0.421 $ &                           $ 83.625 \pm 0.398 $ \\
\midrule
\multirow{4}{*}{indicator}       & DeepSig &  $  \mathbf{ \underline{ 99.960 \pm 0.013 }} $ &  $  \mathbf{ \underline{ 98.068 \pm 0.184 }} $ &  $  \mathbf{ \underline{ 98.056 \pm 0.185 }} $ \\
                                 & RNN     &               $  \mathbf{ 99.955 \pm 0.009 } $ &            $  \underline{ 97.266 \pm 0.439 } $ &            $  \underline{ 97.238 \pm 0.447 } $ \\
                                 & RNNSig  &                           $ 99.747 \pm 0.028 $ &                           $ 93.488 \pm 0.616 $ &                           $ 93.408 \pm 0.613 $ \\
                                 & Sig     &                           $ 99.410 \pm 0.031 $ &                           $ 90.591 \pm 0.306 $ &                           $ 90.492 \pm 0.308 $ \\
\midrule
\multirow{4}{*}{linear}          & DeepSig &                           $ 99.458 \pm 0.052 $ &                           $ 91.567 \pm 0.412 $ &                           $ 91.452 \pm 0.416 $ \\
                                 & RNN     &                           $ 99.489 \pm 0.093 $ &                           $ 91.608 \pm 0.609 $ &                           $ 91.492 \pm 0.608 $ \\
                                 & RNNSig  &                           $ 99.446 \pm 0.039 $ &                           $ 90.259 \pm 0.859 $ &                           $ 90.143 \pm 0.869 $ \\
                                 & Sig     &                           $ 98.963 \pm 0.084 $ &                           $ 87.254 \pm 0.437 $ &                           $ 87.141 \pm 0.458 $ \\
\midrule
\multirow{4}{*}{zero}            & DeepSig &                           $ 99.391 \pm 0.071 $ &                           $ 91.121 \pm 0.406 $ &                           $ 91.012 \pm 0.403 $ \\
                                 & RNN     &                           $ 99.551 \pm 0.031 $ &                           $ 91.765 \pm 0.283 $ &                           $ 91.670 \pm 0.304 $ \\
                                 & RNNSig  &                           $ 99.321 \pm 0.033 $ &                           $ 89.543 \pm 0.412 $ &                           $ 89.457 \pm 0.417 $ \\
                                 & Sig     &                           $ 98.544 \pm 0.069 $ &                           $ 84.269 \pm 0.445 $ &                           $ 84.185 \pm 0.454 $ \\
\bottomrule
\end{tabular}

%% file: tables/repetitions_CharacterTrajectories_Random.tex
\sisetup{
  detect-weight           = true,
  detect-inline-weight    = math,
  separate-uncertainty    = true,
  table-align-uncertainty = true,
  table-text-alignment    = center,
}

\robustify\bfseries

\begin{tabular}{lllll}
\toprule
Imputation                       & Model   &                                        {w-AUROC} &                                            BAC &                                       Accuracy \\
\midrule
\multirow{4}{*}{GP-PoM}          & DeepSig &                           $ 99.698 \pm 0.393 $ &                           $ 94.011 \pm 5.037 $ &                           $ 93.635 \pm 5.335 $ \\
                                 & RNN     &            $  \underline{ 99.970 \pm 0.011 } $ &                           $ 98.011 \pm 0.512 $ &                           $ 98.106 \pm 0.508 $ \\
                                 & RNNSig  &                           $ 99.787 \pm 0.074 $ &                           $ 93.308 \pm 0.960 $ &                           $ 93.844 \pm 0.903 $ \\
                                 & Sig     &                           $ 99.578 \pm 0.031 $ &                           $ 89.570 \pm 0.938 $ &                           $ 89.930 \pm 0.914 $ \\
\midrule
\multirow{4}{*}{GP}              & DeepSig &                           $ 98.994 \pm 1.088 $ &                           $ 90.821 \pm 2.361 $ &                           $ 90.471 \pm 2.347 $ \\
                                 & RNN     &                           $ 99.909 \pm 0.032 $ &                           $ 96.276 \pm 0.691 $ &                           $ 96.492 \pm 0.715 $ \\
                                 & RNNSig  &                           $ 99.400 \pm 0.094 $ &                           $ 87.587 \pm 2.054 $ &                           $ 88.141 \pm 1.959 $ \\
                                 & Sig     &                           $ 94.862 \pm 1.779 $ &                           $ 61.280 \pm 6.440 $ &                           $ 62.446 \pm 6.493 $ \\
\midrule
\multirow{4}{*}{causal}          & DeepSig &                           $ 99.963 \pm 0.023 $ &                           $ 97.774 \pm 0.228 $ &                           $ 97.953 \pm 0.182 $ \\
                                 & RNN     &                           $ 99.953 \pm 0.023 $ &                           $ 97.657 \pm 0.720 $ &                           $ 97.813 \pm 0.676 $ \\
                                 & RNNSig  &                           $ 99.814 \pm 0.044 $ &                           $ 93.268 \pm 0.730 $ &                           $ 93.747 \pm 0.743 $ \\
                                 & Sig     &                           $ 96.736 \pm 0.578 $ &                           $ 71.393 \pm 3.784 $ &                           $ 73.245 \pm 3.642 $ \\
\midrule
\multirow{4}{*}{forward-filling} & DeepSig &                           $ 99.965 \pm 0.030 $ &                           $ 97.974 \pm 0.381 $ &                           $ 98.120 \pm 0.365 $ \\
                                 & RNN     &                           $ 99.954 \pm 0.010 $ &                           $ 97.786 \pm 0.308 $ &                           $ 97.939 \pm 0.281 $ \\
                                 & RNNSig  &                           $ 99.840 \pm 0.047 $ &                           $ 94.110 \pm 0.774 $ &                           $ 94.596 \pm 0.745 $ \\
                                 & Sig     &                           $ 54.308 \pm 4.187 $ &                            $ 7.387 \pm 2.995 $ &                            $ 7.117 \pm 2.417 $ \\
\midrule
\multirow{4}{*}{indicator}       & DeepSig &                           $ 99.955 \pm 0.033 $ &               $  \mathbf{ 98.626 \pm 0.500 } $ &               $  \mathbf{ 98.733 \pm 0.481 } $ \\
                                 & RNN     &                           $ 99.953 \pm 0.024 $ &                           $ 97.502 \pm 0.527 $ &                           $ 97.660 \pm 0.499 $ \\
                                 & RNNSig  &                           $ 99.755 \pm 0.078 $ &                           $ 93.091 \pm 1.056 $ &                           $ 93.635 \pm 0.952 $ \\
                                 & Sig     &                          $ 66.917 \pm 18.306 $ &                          $ 18.481 \pm 18.692 $ &                          $ 19.067 \pm 19.165 $ \\
\midrule
\multirow{4}{*}{linear}          & DeepSig &  $  \mathbf{ \underline{ 99.984 \pm 0.007 }} $ &  $  \mathbf{ \underline{ 98.898 \pm 0.205 }} $ &  $  \mathbf{ \underline{ 98.997 \pm 0.201 }} $ \\
                                 & RNN     &                           $ 99.928 \pm 0.043 $ &                           $ 97.668 \pm 0.897 $ &                           $ 97.786 \pm 0.802 $ \\
                                 & RNNSig  &                           $ 99.767 \pm 0.037 $ &                           $ 92.754 \pm 0.662 $ &                           $ 93.273 \pm 0.656 $ \\
                                 & Sig     &                           $ 55.023 \pm 6.655 $ &                            $ 9.436 \pm 3.349 $ &                            $ 9.958 \pm 4.097 $ \\
\midrule
\multirow{4}{*}{zero}            & DeepSig &               $  \mathbf{ 99.980 \pm 0.013 } $ &            $  \underline{ 98.337 \pm 0.644 } $ &            $  \underline{ 98.454 \pm 0.616 } $ \\
                                 & RNN     &                           $ 99.887 \pm 0.052 $ &                           $ 96.004 \pm 1.074 $ &                           $ 96.253 \pm 1.046 $ \\
                                 & RNNSig  &                           $ 99.685 \pm 0.063 $ &                           $ 92.154 \pm 0.878 $ &                           $ 92.744 \pm 0.820 $ \\
                                 & Sig     &                           $ 96.997 \pm 0.388 $ &                           $ 69.963 \pm 4.208 $ &                           $ 71.699 \pm 4.002 $ \\
\bottomrule
\end{tabular}

%% file: tables/repetitions_PenDigits_Random.tex
\begin{tabular}{lllll}
\toprule
metric &                                        w-AUROC &                                            BAC &                                       Accuracy \\
\midrule
\multirow{4}{*}{GP-PoM}          & DeepSig &                           $ 99.515 \pm 0.078 $ &                           $ 92.151 \pm 0.555 $ &                           $ 92.098 \pm 0.548 $ \\
                                 & RNN     &                           $ 99.564 \pm 0.072 $ &            $  \underline{ 92.757 \pm 0.735 } $ &            $  \underline{ 92.699 \pm 0.733 } $ \\
                                 & RNNSig  &                           $ 98.967 \pm 0.253 $ &                           $ 88.148 \pm 1.588 $ &                           $ 88.113 \pm 1.579 $ \\
                                 & Sig     &                           $ 99.028 \pm 0.099 $ &                           $ 87.352 \pm 0.898 $ &                           $ 87.290 \pm 0.903 $ \\
\midrule
\multirow{4}{*}{GP}              & DeepSig &                           $ 90.509 \pm 0.164 $ &                           $ 54.545 \pm 0.426 $ &                           $ 54.513 \pm 0.451 $ \\
                                 & RNN     &                           $ 91.961 \pm 0.856 $ &                           $ 57.930 \pm 2.079 $ &                           $ 57.900 \pm 2.088 $ \\
                                 & RNNSig  &                           $ 86.740 \pm 0.585 $ &                           $ 46.842 \pm 1.255 $ &                           $ 46.867 \pm 1.218 $ \\
                                 & Sig     &                           $ 83.511 \pm 0.485 $ &                           $ 41.747 \pm 0.428 $ &                           $ 41.809 \pm 0.425 $ \\
\midrule
\multirow{4}{*}{causal}          & DeepSig &                           $ 99.096 \pm 0.116 $ &                           $ 89.480 \pm 0.359 $ &                           $ 89.434 \pm 0.362 $ \\
                                 & RNN     &                           $ 99.288 \pm 0.066 $ &                           $ 89.526 \pm 0.535 $ &                           $ 89.474 \pm 0.539 $ \\
                                 & RNNSig  &                           $ 99.165 \pm 0.067 $ &                           $ 88.807 \pm 0.613 $ &                           $ 88.759 \pm 0.617 $ \\
                                 & Sig     &                           $ 97.870 \pm 0.224 $ &                           $ 80.065 \pm 0.980 $ &                           $ 80.011 \pm 0.971 $ \\
\midrule
\multirow{4}{*}{forward-filling} & DeepSig &                           $ 99.141 \pm 0.068 $ &                           $ 88.974 \pm 0.656 $ &                           $ 88.902 \pm 0.644 $ \\
                                 & RNN     &                           $ 99.311 \pm 0.067 $ &                           $ 90.067 \pm 0.247 $ &                           $ 90.029 \pm 0.247 $ \\
                                 & RNNSig  &                           $ 99.203 \pm 0.063 $ &                           $ 88.930 \pm 0.513 $ &                           $ 88.902 \pm 0.528 $ \\
                                 & Sig     &                           $ 98.425 \pm 0.069 $ &                           $ 84.458 \pm 0.468 $ &                           $ 84.374 \pm 0.477 $ \\
\midrule
\multirow{4}{*}{indicator}       & DeepSig &               $  \mathbf{ 99.607 \pm 0.059 } $ &               $  \mathbf{ 93.156 \pm 0.738 } $ &               $  \mathbf{ 93.087 \pm 0.751 } $ \\
                                 & RNN     &  $  \mathbf{ \underline{ 99.733 \pm 0.044 }} $ &  $  \mathbf{ \underline{ 94.124 \pm 0.412 }} $ &  $  \mathbf{ \underline{ 94.071 \pm 0.415 }} $ \\
                                 & RNNSig  &                           $ 99.549 \pm 0.041 $ &                           $ 91.604 \pm 0.278 $ &                           $ 91.532 \pm 0.268 $ \\
                                 & Sig     &                           $ 98.708 \pm 0.040 $ &                           $ 84.544 \pm 0.538 $ &                           $ 84.505 \pm 0.563 $ \\
\midrule
\multirow{4}{*}{linear}          & DeepSig &                           $ 99.407 \pm 0.151 $ &                           $ 91.418 \pm 1.075 $ &                           $ 91.366 \pm 1.086 $ \\
                                 & RNN     &                           $ 99.510 \pm 0.041 $ &                           $ 91.862 \pm 0.582 $ &                           $ 91.812 \pm 0.594 $ \\
                                 & RNNSig  &            $  \underline{ 99.591 \pm 0.036 } $ &                           $ 91.556 \pm 0.518 $ &                           $ 91.521 \pm 0.539 $ \\
                                 & Sig     &                           $ 99.029 \pm 0.094 $ &                           $ 87.116 \pm 0.612 $ &                           $ 87.038 \pm 0.612 $ \\
\midrule
\multirow{4}{*}{zero}            & DeepSig &                           $ 99.334 \pm 0.077 $ &                           $ 89.774 \pm 0.541 $ &                           $ 89.686 \pm 0.553 $ \\
                                 & RNN     &                           $ 99.403 \pm 0.112 $ &                           $ 90.729 \pm 0.618 $ &                           $ 90.698 \pm 0.620 $ \\
                                 & RNNSig  &                           $ 99.150 \pm 0.046 $ &                           $ 87.948 \pm 0.248 $ &                           $ 87.879 \pm 0.243 $ \\
                                 & Sig     &                           $ 98.623 \pm 0.073 $ &                           $ 83.935 \pm 0.382 $ &                           $ 83.905 \pm 0.375 $ \\
\bottomrule
\end{tabular}

%% file: tables/repetitions_LSST_Label-based.tex
\begin{tabular}{llllll}
\toprule
Imputation                       & Model   &                                       w-AUROC &                                            BAC &                                       Accuracy \\
\midrule
\multirow{4}{*}{GP-PoM}          & DeepSig &                           $ 64.414 \pm 6.770 $ &                           $ 13.857 \pm 2.394 $ &                            $ 8.743 \pm 2.889 $ \\
                                 & RNN     &                           $ 82.808 \pm 4.284 $ &                           $ 29.483 \pm 7.106 $ &                          $ 29.781 \pm 11.520 $ \\
                                 & RNNSig  &                           $ 82.934 \pm 1.495 $ &                           $ 31.751 \pm 2.338 $ &                           $ 35.442 \pm 3.621 $ \\
                                 & Sig     &                           $ 58.820 \pm 1.427 $ &                           $ 13.779 \pm 0.535 $ &                           $ 10.187 \pm 1.418 $ \\
\midrule
\multirow{4}{*}{GP}              & DeepSig &                           $ 59.686 \pm 3.444 $ &                           $ 13.144 \pm 3.924 $ &                           $ 28.333 \pm 5.465 $ \\
                                 & RNN     &                           $ 69.269 \pm 1.469 $ &                           $ 16.558 \pm 0.465 $ &                           $ 33.903 \pm 0.356 $ \\
                                 & RNNSig  &                           $ 60.032 \pm 0.476 $ &                           $ 16.779 \pm 1.559 $ &                           $ 33.627 \pm 0.213 $ \\
                                 & Sig     &                           $ 57.381 \pm 0.555 $ &                           $ 14.062 \pm 1.174 $ &                           $ 24.294 \pm 2.135 $ \\
\midrule
\multirow{4}{*}{causal}          & DeepSig &                           $ 75.336 \pm 3.543 $ &                           $ 25.481 \pm 5.868 $ &                           $ 39.400 \pm 5.824 $ \\
                                 & RNN     &                           $ 84.107 \pm 0.606 $ &                           $ 37.762 \pm 1.736 $ &                           $ 53.009 \pm 1.684 $ \\
                                 & RNNSig  &                           $ 82.343 \pm 0.235 $ &                           $ 33.570 \pm 2.776 $ &                           $ 50.284 \pm 0.973 $ \\
                                 & Sig     &                           $ 58.837 \pm 3.205 $ &                           $ 12.262 \pm 2.841 $ &                           $ 34.161 \pm 1.295 $ \\
\midrule
\multirow{4}{*}{forward-filling} & DeepSig &                           $ 77.758 \pm 2.562 $ &                           $ 27.473 \pm 4.165 $ &                           $ 42.238 \pm 5.435 $ \\
                                 & RNN     &                           $ 84.153 \pm 0.675 $ &                           $ 38.621 \pm 1.618 $ &                           $ 52.976 \pm 0.667 $ \\
                                 & RNNSig  &                           $ 82.430 \pm 0.439 $ &                           $ 34.078 \pm 1.399 $ &                           $ 50.560 \pm 0.678 $ \\
                                 & Sig     &                           $ 64.200 \pm 0.808 $ &                           $ 14.779 \pm 1.184 $ &                           $ 35.255 \pm 0.520 $ \\
\midrule
\multirow{4}{*}{indicator}       & DeepSig &               $  \mathbf{ 95.351 \pm 1.044 } $ &               $  \mathbf{ 52.124 \pm 2.147 } $ &               $  \mathbf{ 77.283 \pm 2.231 } $ \\
                                 & RNN     &  $  \mathbf{ \underline{ 98.132 \pm 0.242 }} $ &  $  \mathbf{ \underline{ 61.893 \pm 3.862 }} $ &  $  \mathbf{ \underline{ 83.609 \pm 1.452 }} $ \\
                                 & RNNSig  &                           $ 82.635 \pm 1.816 $ &                           $ 29.122 \pm 3.102 $ &                           $ 41.152 \pm 2.456 $ \\
                                 & Sig     &                           $ 57.084 \pm 0.863 $ &                           $ 12.806 \pm 0.951 $ &                           $ 32.620 \pm 1.113 $ \\
\midrule
\multirow{4}{*}{linear}          & DeepSig &                           $ 73.965 \pm 3.208 $ &                           $ 21.356 \pm 0.745 $ &                           $ 36.504 \pm 6.349 $ \\
                                 & RNN     &                           $ 84.931 \pm 0.301 $ &                           $ 40.098 \pm 0.968 $ &            $  \underline{ 54.023 \pm 0.822 } $ \\
                                 & RNNSig  &                           $ 82.883 \pm 0.805 $ &                           $ 32.553 \pm 1.276 $ &                           $ 49.327 \pm 2.426 $ \\
                                 & Sig     &                           $ 66.687 \pm 1.812 $ &                           $ 17.340 \pm 0.777 $ &                           $ 35.726 \pm 0.571 $ \\
\midrule
\multirow{4}{*}{zero}            & DeepSig &                           $ 75.783 \pm 1.217 $ &                           $ 23.598 \pm 3.268 $ &                           $ 42.019 \pm 1.718 $ \\
                                 & RNN     &            $  \underline{ 87.908 \pm 0.952 } $ &            $  \underline{ 40.479 \pm 1.890 } $ &                           $ 53.268 \pm 2.066 $ \\
                                 & RNNSig  &                           $ 79.432 \pm 0.496 $ &                           $ 34.381 \pm 1.176 $ &                           $ 46.521 \pm 0.446 $ \\
                                 & Sig     &                           $ 52.972 \pm 1.080 $ &                           $ 11.951 \pm 1.116 $ &                           $ 28.532 \pm 8.378 $ \\
\bottomrule
\end{tabular}

%% file: tables/repetitions_LSST_Random.tex
\begin{tabular}{lllll}
\toprule
Imputation                       & Model   &                                        w-AUROC &                                            BAC &                                       Accuracy \\
\midrule
\multirow{4}{*}{GP-PoM}          & DeepSig &                           $ 70.771 \pm 2.903 $ &                           $ 21.823 \pm 2.656 $ &                           $ 40.114 \pm 2.400 $ \\
                                 & RNN     &                           $ 82.322 \pm 0.833 $ &            $  \underline{ 37.372 \pm 1.945 } $ &                           $ 52.506 \pm 2.559 $ \\
                                 & RNNSig  &                           $ 75.839 \pm 1.576 $ &                           $ 26.680 \pm 1.854 $ &                           $ 41.792 \pm 1.854 $ \\
                                 & Sig     &                           $ 58.799 \pm 0.561 $ &                           $ 13.200 \pm 1.300 $ &                           $ 34.615 \pm 0.802 $ \\
\midrule
\multirow{4}{*}{GP}              & DeepSig &                           $ 62.133 \pm 0.957 $ &                           $ 16.191 \pm 1.807 $ &                           $ 34.324 \pm 0.161 $ \\
                                 & RNN     &                           $ 62.638 \pm 2.952 $ &                           $ 17.363 \pm 2.249 $ &                           $ 34.515 \pm 1.155 $ \\
                                 & RNNSig  &                           $ 60.593 \pm 0.405 $ &                           $ 17.138 \pm 1.515 $ &                           $ 33.710 \pm 0.651 $ \\
                                 & Sig     &                           $ 57.978 \pm 1.329 $ &                           $ 13.614 \pm 1.147 $ &                           $ 33.104 \pm 0.400 $ \\
\midrule
\multirow{4}{*}{causal}          & DeepSig &                           $ 74.266 \pm 2.119 $ &                           $ 22.177 \pm 2.683 $ &                           $ 34.096 \pm 6.644 $ \\
                                 & RNN     &               $  \mathbf{ 83.938 \pm 0.729 } $ &               $  \mathbf{ 37.407 \pm 2.269 } $ &  $  \mathbf{ \underline{ 54.558 \pm 1.074 }} $ \\
                                 & RNNSig  &                           $ 77.195 \pm 5.004 $ &                           $ 31.676 \pm 5.485 $ &                           $ 46.399 \pm 5.323 $ \\
                                 & Sig     &                           $ 52.682 \pm 0.817 $ &                           $ 11.625 \pm 0.808 $ &                           $ 26.026 \pm 9.910 $ \\
\midrule
\multirow{4}{*}{forward-filling} & DeepSig &                           $ 78.760 \pm 1.204 $ &                           $ 27.871 \pm 2.226 $ &                           $ 46.853 \pm 1.426 $ \\
                                 & RNN     &  $  \mathbf{ \underline{ 84.267 \pm 0.570 }} $ &  $  \mathbf{ \underline{ 38.320 \pm 0.546 }} $ &            $  \underline{ 53.236 \pm 1.298 } $ \\
                                 & RNNSig  &                           $ 82.291 \pm 0.348 $ &                           $ 33.517 \pm 1.103 $ &                           $ 50.203 \pm 1.081 $ \\
                                 & Sig     &                           $ 56.031 \pm 2.311 $ &                           $ 12.981 \pm 1.521 $ &                          $ 28.516 \pm 11.770 $ \\
\midrule
\multirow{4}{*}{indicator}       & DeepSig &                           $ 69.863 \pm 1.579 $ &                           $ 20.810 \pm 1.165 $ &                           $ 36.399 \pm 3.986 $ \\
                                 & RNN     &                           $ 76.956 \pm 2.996 $ &                           $ 29.178 \pm 2.580 $ &                           $ 42.182 \pm 4.442 $ \\
                                 & RNNSig  &                           $ 63.700 \pm 1.525 $ &                           $ 18.367 \pm 1.625 $ &                           $ 30.308 \pm 1.881 $ \\
                                 & Sig     &                           $ 53.831 \pm 1.467 $ &                           $ 12.214 \pm 2.317 $ &                           $ 32.668 \pm 0.363 $ \\
\midrule
\multirow{4}{*}{linear}          & DeepSig &                           $ 75.163 \pm 4.039 $ &                           $ 23.657 \pm 4.388 $ &                           $ 40.324 \pm 5.440 $ \\
                                 & RNN     &            $  \underline{ 83.777 \pm 0.512 } $ &                           $ 36.819 \pm 2.375 $ &               $  \mathbf{ 53.439 \pm 0.607 } $ \\
                                 & RNNSig  &                           $ 81.588 \pm 1.000 $ &                           $ 29.814 \pm 1.953 $ &                           $ 49.286 \pm 1.578 $ \\
                                 & Sig     &                           $ 65.499 \pm 3.060 $ &                           $ 17.113 \pm 3.513 $ &                           $ 36.334 \pm 0.979 $ \\
\midrule
\multirow{4}{*}{zero}            & DeepSig &                           $ 69.513 \pm 4.856 $ &                           $ 15.231 \pm 4.668 $ &                           $ 35.272 \pm 4.149 $ \\
                                 & RNN     &                           $ 81.739 \pm 0.416 $ &                           $ 35.580 \pm 2.688 $ &                           $ 50.073 \pm 2.167 $ \\
                                 & RNNSig  &                           $ 77.597 \pm 0.632 $ &                           $ 31.294 \pm 1.615 $ &                           $ 45.953 \pm 1.286 $ \\
                                 & Sig     &                           $ 52.900 \pm 1.487 $ &                           $ 11.870 \pm 2.303 $ &                           $ 32.612 \pm 0.650 $ \\
\bottomrule
\end{tabular}

%% file: tables/repetitions_Physionet2012.tex
\begin{tabular}{llll}
\toprule
Imputation                       & Model   &                                         AUROC &                              Average Precision \\
\midrule
\multirow{4}{*}{GP-PoM}          & DeepSig &                           $ 82.084 \pm 0.836 $ &                           $ 47.858 \pm 1.362 $ \\
                                 & RNN     &                           $ 83.222 \pm 0.570 $ &                           $ 49.263 \pm 1.115 $ \\
                                 & RNNSig  &                           $ 77.879 \pm 1.072 $ &                           $ 40.157 \pm 0.539 $ \\
                                 & Sig     &                           $ 74.388 \pm 2.211 $ &                           $ 33.909 \pm 2.902 $ \\
\midrule
\multirow{4}{*}{GP}              & DeepSig &                           $ 82.164 \pm 0.245 $ &                           $ 47.707 \pm 0.971 $ \\
                                 & RNN     &                           $ 81.196 \pm 0.953 $ &                           $ 47.322 \pm 1.642 $ \\
                                 & RNNSig  &                           $ 74.665 \pm 1.763 $ &                           $ 36.585 \pm 1.549 $ \\
                                 & Sig     &                           $ 70.984 \pm 1.611 $ &                           $ 30.886 \pm 2.115 $ \\
\midrule
\multirow{4}{*}{causal}          & DeepSig &                           $ 83.487 \pm 0.574 $ &                           $ 48.924 \pm 0.935 $ \\
                                 & RNN     &            $  \underline{ 84.689 \pm 0.325 } $ &  $  \mathbf{ \underline{ 52.646 \pm 0.460 }} $ \\
                                 & RNNSig  &                           $ 82.074 \pm 0.080 $ &                           $ 47.691 \pm 0.214 $ \\
                                 & Sig     &                           $ 83.499 \pm 0.597 $ &                           $ 47.809 \pm 0.783 $ \\
\midrule
\multirow{4}{*}{forward-filling} & DeepSig &                           $ 82.766 \pm 0.646 $ &                           $ 47.971 \pm 1.562 $ \\
                                 & RNN     &  $  \mathbf{ \underline{ 84.954 \pm 0.157 }} $ &               $  \mathbf{ 52.427 \pm 0.521 } $ \\
                                 & RNNSig  &                           $ 80.916 \pm 0.607 $ &                           $ 44.979 \pm 1.347 $ \\
                                 & Sig     &                           $ 84.328 \pm 0.213 $ &                           $ 51.043 \pm 0.547 $ \\
\midrule
\multirow{4}{*}{indicator}       & DeepSig &                           $ 82.332 \pm 0.467 $ &                           $ 47.150 \pm 1.020 $ \\
                                 & RNN     &               $  \mathbf{ 84.906 \pm 0.211 } $ &            $  \underline{ 51.887 \pm 0.713 } $ \\
                                 & RNNSig  &                           $ 83.651 \pm 0.199 $ &                           $ 49.872 \pm 0.570 $ \\
                                 & Sig     &                           $ 81.570 \pm 0.473 $ &                           $ 45.807 \pm 1.318 $ \\
\midrule
\multirow{4}{*}{linear}          & DeepSig &                           $ 83.168 \pm 0.650 $ &                           $ 48.937 \pm 1.291 $ \\
                                 & RNN     &                           $ 84.367 \pm 0.176 $ &                           $ 50.431 \pm 0.340 $ \\
                                 & RNNSig  &                           $ 77.336 \pm 0.469 $ &                           $ 41.888 \pm 0.583 $ \\
                                 & Sig     &                           $ 83.354 \pm 0.223 $ &                           $ 48.900 \pm 0.480 $ \\
\midrule
\multirow{4}{*}{zero}            & DeepSig &                           $ 80.101 \pm 2.126 $ &                           $ 44.340 \pm 3.076 $ \\
                                 & RNN     &                           $ 83.571 \pm 0.227 $ &                           $ 49.304 \pm 0.597 $ \\
                                 & RNNSig  &                           $ 76.246 \pm 1.436 $ &                           $ 39.502 \pm 2.232 $ \\
                                 & Sig     &                           $ 80.645 \pm 0.097 $ &                           $ 44.728 \pm 0.264 $ \\
\bottomrule
\end{tabular}